\documentclass{article} 
\usepackage[preprint]{colm2026_conference}

\usepackage[T1]{fontenc}
\usepackage{microtype}
\usepackage{hyperref}
\usepackage{url}
\usepackage{booktabs}
\usepackage{wrapfig}
\usepackage[normalem]{ulem}

\usepackage[utf8]{inputenc}
\usepackage{booktabs}   
\usepackage{makecell}   
\usepackage{multirow}   
\usepackage{graphicx}   
\usepackage{bm}

\usepackage{pifont}
\usepackage{colortbl}

\usepackage{microtype}
\usepackage{amsmath}
\usepackage{amssymb}
\usepackage[dvipsnames]{xcolor}

\usepackage{tikz}
\usetikzlibrary{shapes,decorations}
\usepackage{multicol}
\usepackage{soul}
\def\savelastnode{\pgfextra\edef\tmpA{\tikzlastnode}\endpgfextra}
\def\restorelastnode{\pgfextra\edef\tikzlastnode{\tmpA}\endpgfextra}
\tikzstyle{mybox} = [draw=black, fill=yellow!20, thick,
  rectangle, rounded corners, inner sep=10pt, inner ysep=12pt]
\tikzstyle{fancytitle} =[fill=black, text=white]
\tikzstyle{title} = [append after command={%
  \savelastnode node[fancytitle,right=10pt] at (\tikzlastnode.north west)%
  {\large{#1}}\restorelastnode}]

\tikzstyle{mynewbox} = [draw=black, fill=yellow!20, thick,
  rectangle, rounded corners, inner sep=5pt, inner ysep=3pt]

\newcommand{\smallbluebold}[1]{\textcolor{Blue}{\footnotesize{\textbf{#1}}}}

\makeatother

\usepackage{booktabs}
\usepackage{amsmath}

\usepackage{inconsolata}

\usepackage{graphicx}

\usepackage{lineno}

\definecolor{darkblue}{rgb}{0, 0, 0.5}
\hypersetup{colorlinks=true, citecolor=darkblue, linkcolor=darkblue, urlcolor=darkblue}

\title{Co-Evolving LLM Decision and Skill Bank Agents \\ for Long-Horizon Tasks} 

\author{Xiyang Wu\textsuperscript{1} \quad Zongxia Li\textsuperscript{1} \quad Guangyao Shi\textsuperscript{2} \quad Alexander Duffy\textsuperscript{3} \\ \textbf{Tyler Marques}\textsuperscript{3} \quad \textbf{Matthew Lyle Olson}\textsuperscript{4}  \quad \textbf{Tianyi Zhou}\textsuperscript{5} \quad \textbf{Dinesh Manocha}\textsuperscript{1}\\[0.5ex] \textsuperscript{1}University of Maryland \quad 
\textsuperscript{2}University of Southern California  \quad \textsuperscript{3}Good Start Labs \quad 
\\\textsuperscript{4}Independent Researcher \quad 
\textsuperscript{5}Mohamed bin Zayed University of Artificial Intelligence   \\ {\tt\small wuxiyang@umd.edu \quad zli12321@umd.edu \quad alex@goodstartlabs.com}}

\newcommand{\ours}{COS-PLAY}

\begin{document}

\ifcolmsubmission
\linenumbers
\fi

\maketitle

\begin{abstract}
Long-horizon interactive environments are a testbed for evaluating agents' skill usage abilities.
These environments demand multi-step reasoning, the chaining of
multiple skills over many timesteps, and robust decision-making under
delayed rewards and partial observability.
Games are a good testbed for evaluating agent skill usage in environments.
Large Language Models (LLMs) offer a promising alternative as game-playing agents, but they often struggle with consistent long-horizon decision-making because they lack a mechanism to discover, retain, and reuse structured skills across episodes.
We present \textbf{\ours{}}, a co-evolution framework in which an LLM decision agent retrieves skills from a learnable skill bank to guide action taking, while an agent-managed skill pipeline discovers reusable skills from the agent’s unlabeled rollouts to form a skill bank.
Our framework improves both the decision agent to learn better skill retrieval and action generation, while the skill-bank agent continually extracts, refines, and updates skills together with their contracts.
Experiments across six game environments show that \textbf{\ours{}} with an 8B base model achieves over 25.1\% average reward improvement against four frontier LLM baselines on single-player game benchmarks while remaining competitive on multi-player social reasoning games.

\begin{center}
\raisebox{-0.15\height}{\includegraphics[width=0.03\textwidth]{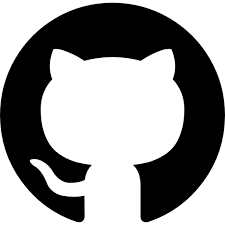}}
Code: \url{https://github.com/wuxiyang1996/COS-PLAY}.
\end{center}    

\end{abstract}

\section{Introduction}
\label{sec:intro}
A central goal of artificial general intelligence is to build autonomous agents that improve through interaction by discovering reusable \emph{skills}, \textit{i.e.}, reusable, temporally extended behavior protocols extracted from trajectories, and invoking them to solve increasingly complex tasks.
This paradigm has a long history, spanning self-improving game agents in Atari and AlphaZero~\citep{mnih2013playing, silver2017mastering} to modern LLM agents for coding, web interaction, and embodied exploration~\citep{wang2025vision, he2025visplay, xiangsystematic, huang2025r}.
%
%
Across these settings, self-learning has expanded from structured games to richer interactive environments, while the core explore--learn--reuse loop remains fundamental to autonomous improvement.
%
%
As agents operate in increasingly complex domains, the challenge shifts from merely acquiring useful skills to organizing, refining, and retrieving them so they can be reused reliably across future tasks.

A natural way to address this challenge is to maintain an external \emph{skill bank}: a structured library of reusable behavior protocols that the agent can acquire, refine, and compose over time~\citep{li2026skillsbench}. 
Recent work increasingly treats \emph{skills} as a distinct abstraction layer for agentic LLMs beyond one-off tool calls or prompts, since skills package reusable procedural knowledge together with execution guidance and applicability constraints~\citep{xu2026agent, jiang2026sok}.
%
Rather than repeatedly retraining the backbone LLM, the system improves by updating this skill bank and learning to invoke the right skills at the right time.
Yet these two components are tightly coupled: a skill bank is only useful if the decision agent can select and execute its skills effectively, and the decision agent is only as capable as the skills available to it.
This interdependence motivates two questions:
\textbf{(Q1)}~Should the skill bank and decision agent be learned \emph{jointly}, rather than separately?
\textbf{(Q2)}~What properties make a skill bank useful for improving decision-making?

To study these questions, we need an environment that reveals co-evolution dynamics: episodes must be long enough to require \emph{multiple} skills in sequence, subgoals must \emph{recur} across episodes so that reuse is meaningful, and rollouts must be cheap enough to support iterative skill-bank refinement. Games offer such a setting in a safe and reproducible sandbox~\citep{wang2023voyager, raad2024scaling, berner2019dota, xu2025agents}, while still demanding long-context understanding, memory, and adaptive decision-making~\citep{hu2024survey}. At the same time, they pose challenges that make naive approaches insufficient: delayed rewards~\citep{liu2025agentic}, compositional strategies, and limited high-quality demonstrations or skill annotations~\citep{liao2025think, lu2025cultivating}. Consequently, many existing game-playing agents still rely on curated human data~\citep{hu2023language} or prolonged training with external feedback, whereas we instead aim to learn a co-evolving skill bank directly from interaction.

In this paper, we propose a \textbf{multi-agent co-evolution} framework for long-horizon game playing. 
The framework comprises: (i)~an LLM-based \textbf{decision agent} that maintains an intention and active skill, retrieves or switches skills as needed, and executes primitive actions conditioned on both; and (ii)~a \textbf{skill bank agent} that performs unsupervised skill discovery and maintenance by segmenting trajectories, learning compact skill contracts, and refining the skill bank over time. 
The two components improve each other in a closed loop: the decision agent uses the current skill bank for multi-step, skill-guided decision-making, while the skill bank agent converts newly collected rollouts into better reusable skills. 
We optimize both agents with Group Relative Policy Optimization (GRPO)~\citep{feng2025group}: the decision agent is trained to improve skill retrieval and action execution, while the skill bank agent is trained for skill segmentation, contract learning, and bank curation.
To our knowledge, this is one of the first frameworks to couple LLM-based game decision-making with an agentic skill-bank pipeline for unsupervised skill discovery and continual refinement in a unified co-evolution loop.
Our main contributions are:
\begin{itemize}
\item  We propose a co-evolution framework for long-horizon gameplay that closes the loop between decision making and skill learning: an LLM-based decision agent interacts with the environment to collect trajectories, and a skill bank agent converts these rollouts into reusable skills that improve future decision making; 
\item We provide a comprehensive evaluation of \ours{} across six game environments requiring multi-hop skill usage, covering both task performance and skill reusability. Built on an 8B base model, \ours{} achieves over \textbf{25.1\%} average gain against four frontier LLM baselines on single-player games while remaining competitive with SOTA LLMs on multi-player social reasoning tasks.
\end{itemize}

\section{Related Work}
\label{sec:related_work}

\paragraph{Agents for Game Playing and Benchmarks.}
Game agents and benchmarks are useful for studying long-horizon decision-making because they stress memory, planning, and real-time action under delayed rewards and complex dynamics.
However, despite rapid progress, relatively few works study game-playing agents at scale due to the difficulty of interactive environments and evaluation.
BALROG~\citep{paglieri2024balrog}, VS-Bench~\citep{xu2025vs}, VideoGameBench~\citep{zhang2025videogamebench}, and VisGym~\citep{wang2026visgym} introduce increasingly challenging benchmarks for long-horizon reasoning, strategic interaction, and multi-step visual decision-making, revealing large gaps between frontier models and human performance.
On the method side, Optimus-1~\citep{li2024optimus}, VARP~\citep{chen2024can}, and AVA~\citep{ma2025vlms} develop agents for Minecraft, ARPGs, and StarCraft II, highlighting the importance of skill exploration, memory, and planning in complex gameplay~\citep{zhai2025agentevolver}.
These advances motivate our framework, which uses games as a controlled setting for co-evolving decision policies with a refineable skill bank. 
Rather than focusing only on benchmarks or game-specific agents, we study how reusable skills can be extracted from trajectories, refined over time, and fed back to improve long-horizon decision-making across diverse games.

\paragraph{Memory and Skill-Augmented Self-Improving Agents.}
Memory- and skill-augmented agents have gained attention because long-horizon decision-making benefits from reusing past experience and executable skills instead of solving each step from scratch, especially in complex multimodal tasks.
Recent work improves self-evolving agents through memory and skill reuse~\citep{li2026skillsbench,li2026mm, huang2025r,li2025self}. PolySkill~\citep{yu2025polyskill} improves skill transfer by decoupling abstract goals from concrete implementations. SAGE~\citep{wang2025reinforcement}, SkillRL~\citep{xia2026skillrl}, SCALAR~\citep{zabounidis2026scalar}, and XSkill~\citep{jiang2026xskill} use skill-augmented learning to improve skill generation, grounding, composition, retrieval, and policy performance. Memweaver~\citep{yu2025memweaver} introduces a hierarchical dual-memory framework combining graph-based retrieval with LLM-summarized cognitive memory. ProcMEM~\citep{mi2026procmem} and CASCADE~\citep{huang2025cascade} emphasize procedural memory and cumulative skill creation. UI-Mem~\citep{xiao2026ui} studies hierarchical experience memory for long-horizon online RL, while MemRL~\citep{zhang2026memrl} and MemSkill~\citep{zhang2026memskill} treat memory retrieval and operations as runtime mechanisms for improving frozen agents.
These works motivate our framework mostly focus on memory augmentation, skill transfer, or runtime retrieval within largely fixed pipelines.
In contrast, our method studies co-evolution, where a decision agent and skill bank improve each other over time by automatically extracting, refining, and reusing skills from unlabeled trajectories to improve downstream decision-making.
\section{Problem Formulation}
\label{sec:problem}

\textbf{Preliminary.} We consider an interactive game environment $\mathcal{E} = (\mathcal{O}, \mathcal{A}, P, R, \gamma, T)$ with horizon $T$, where $\mathcal{O}$ is the observation space, $\mathcal{A}$ is the action space, $P(o_{t+1}\mid o_t,a_t)$ denotes the environment transition dynamics, $R(o_t,a_t,o_{t+1}) \in \mathbb{R}$ is the reward function, and $\gamma \in (0,1]$ is the discount factor. At each timestep $t$, an agent observes $o_t \in \mathcal{O}$, takes an action $a_t \in \mathcal{A}$, receives reward $r_t \in \mathbb{R}$, and transitions to the next observation $o_{t+1}$.
A trajectory (episode) is denoted by $\tau = (e_1, e_2, \dots, e_T)$, where each step is represented as an experience
\begin{align}
    e_t = (o_t, a_t, r_t, o_{t+1}, d_t, z_t) \notag
\end{align}
Here $d_t \in \{0,1\}$ is the termination flag, and $z_t$ is the agent's latent intention state at timestep $t$, representing its internal strategic interpretation of the environment~\citep{qi2018intent, wu2023intent}.
This latent intention may correspond to either a short-horizon action intention for the next step or a higher-level strategic skill that guides behavior over the next several steps. In our framework, $z_t$ is generated from the agent's understanding of the current environment state and is iteratively updated throughout the trajectory as the agent interacts with the environment and refines its strategic objective.

\noindent \textbf{Skills.}
A skill \(s_k\) is a reusable, temporally extended behavior abstraction extracted from trajectory segments. 
Each skill is stored in the skill bank as a structured \emph{skill protocol} with the following components: \textit{Summary}, which describes the skill’s purpose; \textit{Pre-condition}, which specifies when the skill is applicable; \textit{Plan}, which outlines how the skill should be carried out; \textit{Success/Abort Criteria}, which indicate when execution should terminate successfully or be abandoned; and \textit{Contract}, which summarizes the state changes the skill is expected to produce. 
These components form the skill’s externally queryable interface for retrieval and execution, while the skill itself remains grounded in the set of supporting trajectory segments from which the protocol and contract are learned and refined over time.



\noindent \textbf{Refineable Skill Bank.}
A key property of our framework is that skills are not fixed after initialization. Instead, as interactive learning between the agents and the environment continues, the skill bank is continuously updated and refined with new trajectory evidence.
As new rollouts are collected, the skill bank agent updates the bank by segmenting trajectories, refining existing skills, and storing emerging skills.

\noindent \textbf{Skill-augmented Decision-making.}
Given the current skill bank \(\mathcal{B}\), the decision agent maintains an intention state \(z_t\) that captures its current strategic focus and skill-level subgoal. At each timestep, it first retrieves candidate skills from the bank, selects one based on the current observation and intention, updates its intention conditioned on the selected skill, and finally executes an action. Formally,
\begin{align}
    \tilde{s}_t &= \pi_{\theta}^{\mathrm{skill}}(o_t, \mathcal{B}), \\
    z_t &= \pi_{\theta}^{\mathrm{int}}(o_t, \tilde{s}_t), \\
    a_t &\sim \pi_{\theta}^{\mathrm{act}}(\cdot \mid o_t, z_t, \tilde{s}_t).
\end{align}
We model the decision agent with three components serving distinct roles: \(\pi_{\theta}^{\mathrm{skill}}\) for skill retrieval-and-selection, \(\pi_{\theta}^{\mathrm{int}}\) for intention updating, and \(\pi_{\theta}^{\mathrm{act}}\) for action execution.
Here, \(\tilde{s}_t\) denotes the retrieved skill or skill set at timestep \(t\), and \(z_t\) denotes the updated intention state conditioned on the current observation and retrieved skill. The agent is trained to maximize the expected cumulative reward:
\begin{align}
\max_{\theta}\; J(\theta)
= \mathbb{E}_{\tau \sim \pi_{\theta}}
\left[\sum_{t=1}^{T} r_t\right].
\end{align}
%


\noindent \textbf{Skill Bank Update.}
We formulate skill-bank refinement iteratively because the skill bank and decision agent co-evolve over training: each round of newly collected trajectories reflects the current policy, while the updated bank shapes subsequent skill retrieval and action execution.
The trajectory set \(\mathcal{D}^{(u+1)}\) collected up to co-evolution iteration \(u+1\) is passed to the skill bank agent to refine the current skill bank \(\mathcal{B}^{(u)}\):
\begin{align}
\mathcal{B}^{(u+1)}=\Phi_{\mathrm{S}}\!\left(\mathcal{B}^{(u)}, \mathcal{D}^{(u+1)}\right),
\end{align}
where \(\mathcal{B}^{(u)}\) and \(\mathcal{B}^{(u+1)}\) denote the skill bank before and after the update, \(\mathcal{D}^{(u+1)}\) denotes the accumulated rollout set, and \(\Phi_{\mathrm{S}}\) denotes the skill-bank update pipeline. 
%
This pipeline continually updates the skill bank so that retrieved skills remain reusable and aligned with the evolving decision policy.
\begin{figure*}[t]
    \centering
    \includegraphics[width=\textwidth]{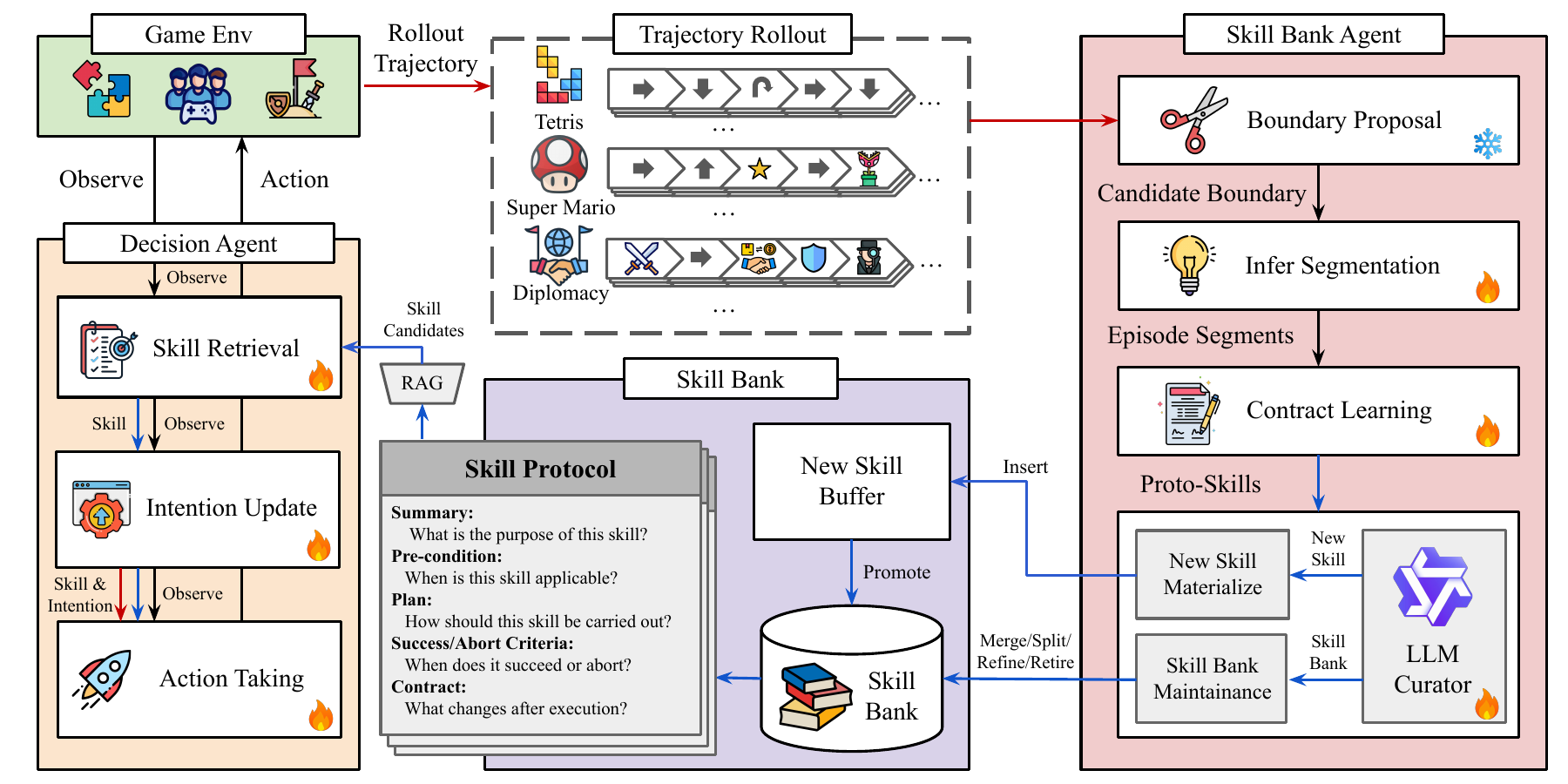}
     \vspace{-10pt}
    \caption{\textbf{Overview of \ours{}.} \ours{} is a multi-agent co-evolution framework that couples gameplay with skill learning. It consists of a decision agent (\textcolor{orange}{\textbf{Orange Box}}), a skill bank agent (\textcolor{red}{\textbf{Red Box}}), and a skill bank (\textcolor{violet}{\textbf{Purple Box}}). The decision agent interacts with the game by retrieving skills, updating intentions, and selecting actions. After each episode, the skill bank agent segments trajectories, learns skill contracts, adds new skills, and updates existing ones through refinement, merging, splitting, or retirement. This loop enables the agent to acquire reusable skills from prior experience and continually improve its policy.}
    \label{fig:teaser}
    \vspace{-12pt}
\end{figure*}

\section{Methodology}
\label{sec:methodology}

We propose a co-evolving multi-agent framework for long-horizon video-game decision-making via unsupervised trajectory decomposition and skill-bank refinement (Figure~\ref{fig:teaser}). The framework has two components:
\textbf{(a) Decision Agent \(A_D\)}: an LLM-based agent that interacts with the game through primitive actions and skill retrieval. At each step, it summarizes the current state, retrieves relevant skill candidates from the skill bank, updates its intention, selects or switches skills when needed, and executes an action.
\textbf{(b) Skill Bank Agent \(A_S\)}: an LLM-based pipeline that converts unlabeled trajectories into reusable protocol-based skills and learns compact effect contracts for them. It updates the skill bank by proposing new skill candidates, refining low-quality skills, and revising skill protocols over time.
Together, the decision agent generates trajectories, and the skill bank agent transforms them into structured skills that support future decisions through skill retrieval and selection.

\subsection{Skill-Augmented Decision Agent}
\label{sec:methodology:decision_agent}

\noindent
%
We design a skill-augmented decision agent that unifies skill retrieval, intention updating, action execution, and reward-based learning for long-horizon gameplay. 
At each step, it maintains a state summary, intention, and active skill; either continues the current skill or retrieves a new one; and executes a primitive action conditioned on the observation, intention, and skill plan. 
The resulting observation and reward are used to update the agent state, while training is driven by a composite reward combining environment feedback, skill-following shaping, and switching cost.

\noindent \textbf{Skill Retrieval and Intention Update.}
The decision agent maintains an active skill and a current intention. Given the current state summary, it either continues the active skill or retrieves a new one from the skill bank when the current skill is unavailable, exhausted, or ineffective. Each skill is a structured context prompt containing protocol steps, preconditions, and termination cues. After each primitive action, the agent updates a short natural-language intention tag summarizing its immediate subgoal. When this intention shifts sharply, it signals a subtask transition and triggers skill switching~\citep{konidaris2010constructing}.

\noindent \textbf{Action Execution.}
At each timestep, the agent executes a primitive environment action conditioned on the current state summary, task context, recent interaction history, the current intention, and, when available, the active skill plan. The active skill does not replace low-level control, but instead guides action generation toward coherent multi-step behavior. The executed action produces the next observation and reward, which are then used to update the agent state and intention.


\subsection{Skill Bank Agent for Skill Discovery and Maintenance}
\label{sec:methodology:skill_bank_agent}
\begin{figure*}[t]
\centering
\small
\setlength{\tabcolsep}{3pt}
\renewcommand{\arraystretch}{1.05}

\begin{tikzpicture}
\node [draw=black, fill=yellow!8, thick, rectangle, rounded corners, inner sep=8pt, inner ysep=8pt] (box){%
\begin{minipage}{0.96\textwidth}
\small\raggedright

\noindent\colorbox{Blue!12}{\parbox{\dimexpr\linewidth-2\fboxsep}{\centering\small\textbf{Skill Discovery Pipeline --- Diplomacy (Austria), one episode}}}
\\[5pt]

\noindent\colorbox{black!8}{\parbox{\dimexpr\linewidth-2\fboxsep}{\small\textbf{(a) Raw Trajectory}}}
\\[1pt]
{\footnotesize\textit{20-step rollout. Shaded rows mark detected transitions.}}
\\[2pt]

\begin{center}
\footnotesize
\begin{tabular}{@{}c l l l c@{}}
\toprule
$t$ & \textbf{State Predicates} & \textbf{Action} & \textbf{Intention} & $r_t$ \\
\midrule
0 & \texttt{3SC, nbr\_intent=\{?\,?\,?\}} & A BUD$\to$TRI & ``scout Italian border'' & 0 \\
1 & \texttt{italy\_in\_VEN (neutral)} & F TRI$\to$ALB & ``secure Adriatic'' & 0 \\
\rowcolor{black!4}
4 & \texttt{russia\_in\_GAL (hostile)} & A VIE$\to$GAL & ``block Russia'' & 0 \\
\cmidrule{1-5}
\rowcolor{orange!10}
5 & \texttt{galicia\_held, serbia\_open} & A BUD$\to$SER & ``take Serbia'' & +1 \\
6 & \texttt{4SC, turkey\_passive} & F ALB$\to$GRE & ``push south'' & +1 \\
\rowcolor{black!4}
9 & \texttt{5SC, italy\_allied} & F GRE$\to$ION & ``enter Mediterranean'' & +1 \\
\cmidrule{1-5}
\rowcolor{red!8}
10 & \texttt{turkey\_attacks\_BUL} & A BUL\,S\,A SER & ``hold Bulgaria'' & 0 \\
11 & \texttt{5SC, east\_front\_stable} & A SER\,S\,A BUL & ``consolidate east'' & 0 \\
\bottomrule
\end{tabular}
\end{center}
\vspace{1pt}

\noindent\colorbox{black!8}{\parbox{\dimexpr\linewidth-2\fboxsep}{\small\textbf{(b) Boundary Proposal}}}
\\[1pt]
{\footnotesize\textit{Three candidates scored; $t{=}3$ falls below threshold.}}
\\[2pt]
\begin{center}
\footnotesize
\begin{tabular}{@{}c c l@{}}
\toprule
\textbf{Candidate} & \textbf{Score} & \textbf{Signals} \\
\midrule
\rowcolor{orange!10}
$t{=}5$ & 0.83 & intention flip (``scout'' $\to$ ``take'') + predicate flip (\texttt{+serbia\_open}) \\
\rowcolor{red!8}
$t{=}10$ & 0.76 & reward drop ($+1 \to 0$) + predicate flip (\texttt{+turkey\_attacks\_BUL}) \\
$t{=}3$ & 0.41 & \sout{minor intention reword} \quad$\leftarrow$ discarded \\
\bottomrule
\end{tabular}
\end{center}
\vspace{1pt}

\noindent
\begin{minipage}[t]{0.45\linewidth}

\noindent\colorbox{black!8}{\parbox{\dimexpr\linewidth-2\fboxsep}{\small\textbf{(c) Infer Segmentation}}}
\\[1pt]
{\footnotesize\textit{Four segments; three match existing bank skills.}}
\\[2pt]
\footnotesize
\begin{tabular}{@{}l l c@{}}
\toprule
\textbf{Segment} & \textbf{Skill Label} & \textbf{Conf.} \\
\midrule
$[0\text{--}4]$ & \textsc{Explore} \textcolor{gray}{(matched)} & 0.87 \\
$[5\text{--}9]$ & \textsc{Setup} \textcolor{gray}{(matched)} & 0.82 \\
$[10\text{--}16]$ & \textsc{Defend} \textcolor{gray}{(matched)} & 0.79 \\
$[17\text{--}19]$ & \emph{new skill} \textcolor{gray}{(none)} & --- \\
\bottomrule
\end{tabular}

\vspace{3pt}

\noindent\colorbox{black!8}{\parbox{\dimexpr\linewidth-2\fboxsep}{\small\textbf{(d) Contract Learning (\textsc{Explore})}}}
\\[1pt]
{\footnotesize\textit{Consensus effects over 28 episodes; one sporadic effect dropped.}}
\\[2pt]
\footnotesize
\textbf{Add:} \texttt{+border\_info(IT)}, \texttt{+border\_info(RU)},\\
\phantom{\textbf{Add:}} \texttt{+gal\_status\_known} \\
\textbf{Del:} \texttt{-unk\_intent(IT)}, \texttt{-unk\_intent(RU)} \\
\textbf{Verify:} 26/28 pass $\to$ rate\,=\,0.93\;$\checkmark$ \\[1pt]
\textcolor{gray}{\texttt{+fleet\_adriatic} (5/28) dropped.}
\end{minipage}
\hfill
\begin{minipage}[t]{0.53\linewidth}

\noindent\colorbox{black!8}{\parbox{\dimexpr\linewidth-2\fboxsep}{\small\textbf{(e) Skill Bank Maintenance}}}
\\[1pt]
{\footnotesize\textit{Bank mutations triggered by this episode.}}
\\[2pt]
\footnotesize
\begin{tabular}{@{}l p{0.82\linewidth}@{}}
\textbf{Refine} & \textsc{Explore}: +1 clause \texttt{gal\_status} \\[1pt]
\textbf{Mat.} & $[17\text{--}19]$ buffered; 12 instances $\to$ \textsc{Negotiate} \\[1pt]
\textbf{Merge} & \textsc{Press.\_Ser.}$\,\approx\,$\textsc{Balk.\_Push} $\to$ \textsc{Setup} \\[1pt]
\textbf{Retire} & \textsc{Scatter\_Units}: 0 uses $\to$ removed \\
\end{tabular}

\vspace{4pt}

\noindent\colorbox{Blue!10}{\parbox{\dimexpr\linewidth-2\fboxsep}{\small\textbf{Stored Skill Entry: \textsc{Explore} \textcolor{gray}{(v2, 28 inst.)}}}}
\\[1pt]
{\footnotesize\textit{The protocol the decision agent queries.}}
\\[2pt]
\footnotesize
\begin{tabular}{@{}r@{\;:\;}p{0.84\linewidth}@{}}
\textit{Summary} & Scout borders, probe neighbor intentions \\
\textit{Pre-cond} & Game start or awareness lost \\
\textit{Plan} & Observe $\to$ Assess $\to$ Commit \\
\textit{Success} & $\geq$3 neighbor intents classified \\
\textit{Abort} & Attack on home centers \\
\textit{Contract} & $+$\texttt{border\_info}($N$),\; $-$\texttt{unk\_intent}($N$) \\
\end{tabular}
\end{minipage}

\end{minipage}
};
\end{tikzpicture}

\vspace{-6pt}
\caption{\textbf{Skill bank agent pipeline on one Diplomacy episode (Austria).} \textbf{(a) Raw Trajectory.}~A decision-agent rollout; shaded rows mark skill transitions. \textbf{(b) Boundary Proposal.}~We score each timestep for transition signals and discard low scorers. \textbf{(c) Infer Segmentation.}~We select true boundaries and label each segment with a bank skill or \emph{new skill}. \textbf{(d) Contract Learning.}~We aggregate state deltas across all instances of \textsc{Explore} and drop sporadic effects to obtain a verified contract. \textbf{(e) Skill Bank Maintenance.}~Four operations (refine, materialize, merge, retire) update the bank. The bottom-right box shows the stored protocol the decision agent queries.}
\label{fig:skill_pipeline_example}
\vspace{-15pt}
\end{figure*}
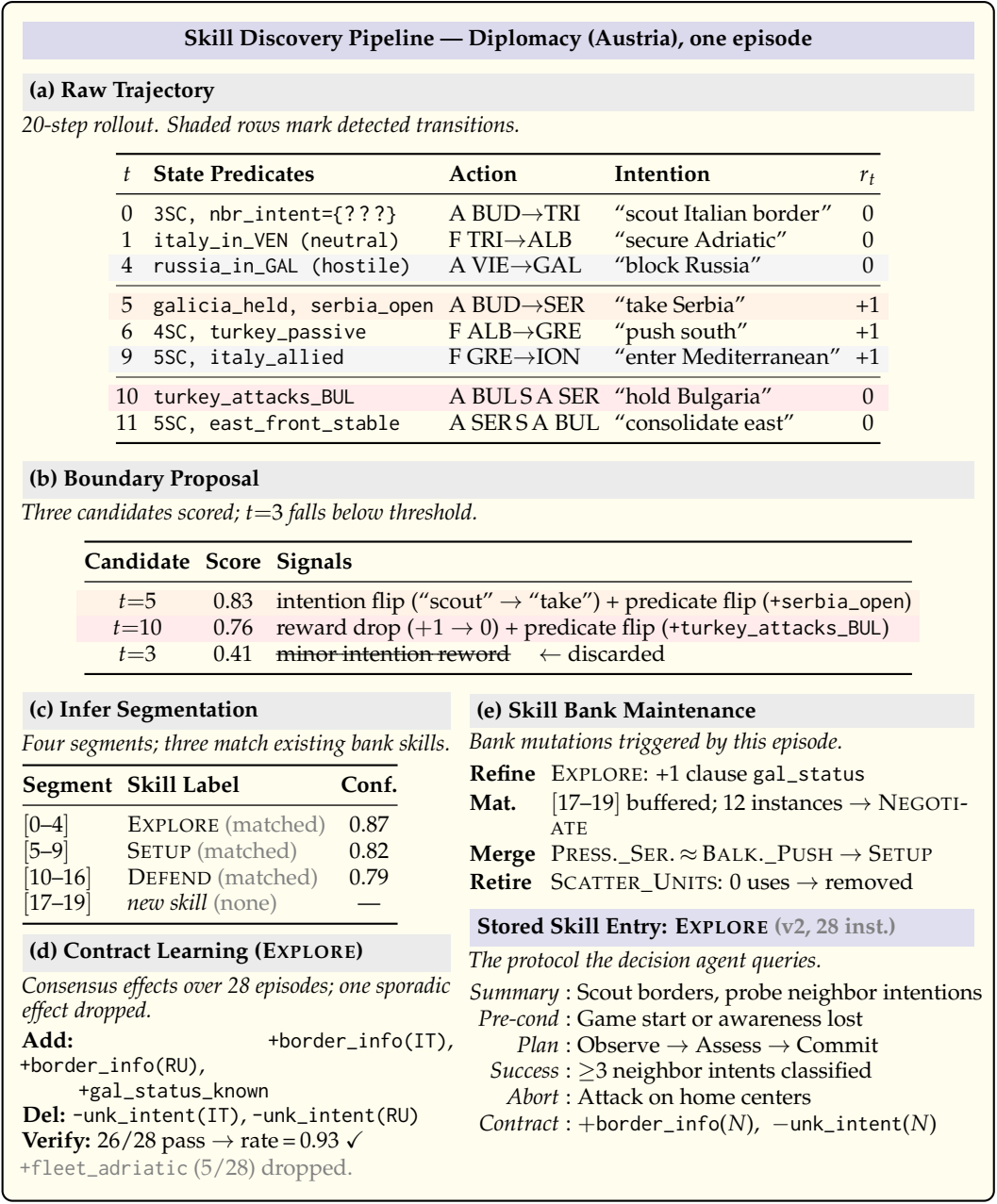

With decision agent rollouts, a skill bank agent analyzes the trajectory to extract reusable skills and maintain the skill bank over time. This agent has a four-stage pipeline: proposing candidate skill boundaries, inferring a segmentation with skill labels, learning and verifying effect contracts, and finally updating the bank through refinement, materialization, merging, splitting, and retirement. In this way, the bank evolves from repeated trajectory evidence while remaining compact, stable, and useful for downstream decision making. Figure~\ref{fig:skill_pipeline_example} walks through the full pipeline on a single Diplomacy episode.



\noindent\textbf{Boundary Proposal.}
Boundary proposal is a heuristic candidate-generation step that identifies plausible skill-transition points in a trajectory.
For each timestep, we compute a boundary score from lightweight local signals that often indicate a change of skill, including predicate flips between adjacent states, intention-tag changes, reward or event spikes, optional surprisal peaks, and transitions between primitive-action execution and skill-selection steps.
Timesteps with high scores are retained as candidate cut points, after which nearby candidates are merged to remove redundancy.
This produces a compact, high-recall boundary set that is later disambiguated by the segmentation module.

\noindent\textbf{Infer Segmentation.}
Starting from the candidate boundaries, we choose the subset that best explains the trajectory as a sequence of skill segments.
Each segment is summarized by its observed effects and compared against skills in the current bank. We first score candidate skills by overlap between the segment’s added and deleted predicates and each skill’s learned \emph{effect contract}, \textit{i.e.}, a compact specification of the state changes that the skill reliably produces when executed successfully, with a small bonus for stronger prior support, and then re-rank the top candidates using the skill bank agent.
If a segment matches an existing skill with high confidence, we assign that label. Otherwise, we mark it as \emph{new skill}, indicating behavior not yet represented in the bank. Its observed effects are kept as provisional evidence, and a reusable effect contract is learned later by aggregating and verifying multiple such segments in the contract learning stage.



\noindent\textbf{Contract Learning.}
For each skill, we aggregate the added and deleted predicates across its decoded segments to learn an \emph{effect contract} that captures its reliable state changes.
We keep only consensus effects that appear consistently across instances, treating infrequent ones as noise.
When enough evidence is available, the contract can be enriched by an LLM-based summarization module, whose suggestions are added to the statistical consensus.
We then verify whether the contract effects are supported by observed state changes, and only contracts with sufficiently high pass rates are written back into the skill bank.
These verified contracts also provide a compatibility prior for segment decoding, giving higher confidence to segments whose observed effects better match a skill’s contract.


\textbf{Skill Bank Maintenance.}
The final stage mutates the skill bank through five operations: \emph{refine}, \emph{materialize}, \emph{merge}, \emph{split}, and \emph{retire}.
Segments labeled as \emph{new skill} are first evaluated for consistency and reuse potential rather than added immediately; promising ones enter a new-skill buffer, where similar candidates are grouped until sufficient evidence supports materializing a new bank entry.
For existing skills, \emph{refine} updates contracts when new verified evidence meaningfully changes their reliable effects, \emph{merge} removes redundant near-duplicate skills with highly overlapping contracts, and \emph{split} marks overly broad or inconsistent skills for later re-segmentation.
We also \emph{retire} skills that are no longer used or supported by recent trajectories, keeping the bank compact and relevant.
The skill bank agent may approve or reject proposed mutations before they are committed to the bank.

\subsection{Co-Evolution Framework}
\label{sec:methodology:co-evolution}
We adopt a \emph{co-evolution} framework in \ours{} where the decision agent and skill bank agent are mutually dependent. The decision agent uses the current skill bank to interact with the environment and collect trajectories; the skill bank agent then segments these into reusable skills, infers contracts, and refines the bank. This closed loop is self-reinforcing: better skills improve decision making, and better rollouts improve skill learning.
Both agents are updated via GRPO \citep{shao2024deepseekmath} using separate LoRA adapters \citep{hu2022lora}. The decision agent has two adapters: \emph{action-taking}, which rewards progress on the active skill while discouraging unnecessary retrieval, and \emph{skill-retrieval}, which is evaluated at episode end and rewards skills that led to useful, contract-satisfying outcomes. This decomposition lets the agent jointly improve \emph{what} skill to retrieve and choose and \emph{how} to execute it.
The skill bank agent uses three adapters. The \emph{segmentation} adapter maps high-value trajectory segments to existing skills while allowing new skills to appear. The \emph{contract} adapter learns compact state-transition summaries that generalize across episodes. The \emph{curator} adapter updates the bank through retention, exploration, and evidence-based merging, splitting, or rejection. Full reward definitions are given in Appendix~\ref{app:rwd_design}.

\section{Experiments}
\label{sec:experiments}


\textbf{Evaluations. } We evaluate on six game environments spanning single-player puzzle and interactive control tasks, including 2048, Candy Crush, Tetris, and Super Mario Bros.~\citep{hu2025lmgame, park2025orak}, as well as multi-player social reasoning games, Avalon and Diplomacy~\citep{zhai2025agentevolver}. Avalon is a team-based hidden-role game in which only one side can win. It is especially challenging for the Good side, which must infer hidden roles from sparse signals such as proposals, votes, and quest outcomes, while Evil players begin fully coordinated and can strategically conceal or sabotage~\citep{light2023avalonbench}. Diplomacy further increases the difficulty by requiring long-context reasoning, negotiation, alliance tracking, and multi-phase planning. Smaller LLMs have recently been shown to struggle with both Avalon and Diplomacy due to their highly social interactive nature, making them an excellent testbed for \ours{} \citep{rahimirad2025bayesian,duffy2026democratizing}. 
%

Across all environments, observations are converted into structured textual states, and the agent interacts through discrete text actions under a unified Gym-style API~\citep{brockman2016openai}. Full environment details, including observations, actions, horizons, and rewards, are provided in Appendix~\ref{app:game_envs}.

\textbf{Cold Start Initialization.}
We use GPT-5.4 as a teacher model to generate 60 seed trajectories per game. We then apply supervised fine-tuning (SFT) on these trajectories to train a Qwen3-8B~\citep{yang2025qwen3} model, which serves as the shared initialization for both the decision agent and the skill bank agent.


\textbf{Co-Evolving Training Setting.} 
The co-evolving training proceeds in three stages: (1) the decision agent interacts with the environment to collect rollouts; (2) the skill-bank agent segments skill candidates from these rollouts and updates the skill bank; and (3) GRPO updates both agents. 
%
Starting from the cold-start model, we attach a separate LoRA
adapters~\citep{hu2022lora} for each trainable function: two for the decision
agent (skill retrieval and action taking) and three for the skill bank agent
(infer segmentation, contract learning, and skill bank maintenance). We discuss
this multi-adapter design further in Appendix~\ref{app:3-in-1}.
For social reasoning games (Avalon and Diplomacy), we use GPT-5-mini~\citep{openai_gpt5mini_model_2026} as opponents to provide strong supervision signals during evolution. Full training details and training curves are in Appendix~\ref{app:key_hparams} and~\ref{app:training_curve}.

\textbf{Reward Design and Metrics.} For single-player games, we report the native game rewards provided by the benchmark environments~\citep{hu2025lmgame,park2025orak}. 
For social games, we use team win rate for Avalon and the number of occupied supply centers for Diplomacy as the main evaluation metrics, following~\citep{zhai2025agentevolver}, to provide a consistent basis for comparison across baselines. More details about reward design are included in Appendix~\ref{app:rwd_design}.

\subsection{Main Results} 
We compare \ours{} with strong frontier LLMs, including \textsc{GPT-5.4}~\citep{openai2026gpt54}, \textsc{Gemini-3.1-Pro}~\citep{googledeepmind2026gemini31pro}, \textsc{Claude-4.6-Sonnet}~\citep{anthropic2026claude46}, and \textsc{gpt-oss-120b}~\citep{openai2025gptoss}. Results are shown in Table~\ref{tab:game_category_results}. General reasoning and qualitative results appear in Appendix~\ref{app:general_reasoning} and~\ref{app:qualitative_analysis}.

\paragraph{\ours{} matches or exceeds frontier LLMs with efficient, few-shot adaptation.}
Overall, \ours{} achieves a substantial average improvement of 25.1\% over \textsc{GPT-5.4} on single-player games over 16 runs. 
More importantly, \ours{} requires only few-shot adaptation to transfer to new benchmarks: each game needs at most 25 iterations to reach strong performance, compared with the hundreds of training steps often required by prior RL-based game agents~\citep{shu2021experience}, suggesting the strong data efficiency, learnability, and cross-domain adaptability of our framework. 

\textbf{A small model with structured state tracking and reusable skills can approach frontier LLMs on social reasoning tasks that require long-context memory and multi-turn consistency.}
For multi-player social games, we use \textsc{GPT-5.4} as the opponent model, against which most non-GPT baselines underperform (Table~\ref{tab:game_category_results}). We provide role-wise and power-wise breakdowns in Table~\ref{tab:avalon_per_role_winrate_compare} and Table~\ref{tab:diplomacy_per_power_mean_sc_ci} in Appendix~\ref{app:more_results}.
In Avalon, \ours{} is comparable to \textsc{Gemini-3.1-Pro} and \textsc{gpt-oss-120b}, trailing by only 1\% in win rate, while in Diplomacy it outperforms \textsc{Gemini-3.1-Pro} by 8.8\%.
These results suggest that structured state tracking and reusable skills provide an effective inductive bias for long-horizon social reasoning, even with a much smaller backbone model.

\subsection{Ablation Study} 
We compare our full pipeline against several variants in Table~\ref{tab:game_category_results}: the base \textsc{Qwen3-8B}~\citep{yang2025qwen3}; the SFT decision agent without a skill bank (\textsc{SFT w/o Skill}); the SFT model with the first skill bank from the initial co-evolution iteration (\textsc{SFT + 1st Skill}-- No Co-Evolution); the SFT model with the final skill bank (\textsc{SFT + Final Skill}-- No Co-Evolution); the GRPO decision agent without a skill bank (\textsc{GRPO w/o Skill}); and the GRPO decision agent with the first skill bank (\textsc{GRPO + 1st Skill}).
The results show that no single component is sufficient. While some variants improve specific games, their gains are inconsistent across domains. \textsc{SFT w/o Skill} improves action formatting but lacks reusable long-horizon structure, whereas \textsc{GRPO w/o Skill} improves behavior yet remains unstable under sparse rewards. Variants with mismatched skill banks perform worse, as the policy and skill bank are optimized for different state distributions, making retrieved skills less aligned and often harmful.
These findings suggest that the main advantage comes not from skills or RL alone, but from co-evolution, which keeps the skill bank aligned with the policy’s changing behavior and task distribution.

\begin{table*}[t]
    \begin{center}
    \resizebox{\textwidth}{!}{
      \begin{tabular}{lcccc c|cc}
      \toprule
          \multicolumn{1}{c}{}
          & \multicolumn{5}{c}{\makecell{\textbf{Single-Player}}}
          & \multicolumn{2}{c}{\makecell{\textbf{Multi-Player}}} \\
          \cmidrule(lr){2-6}\cmidrule(lr){7-8}
          \makecell{\textbf{Model}}
          & \makecell{\textbf{2048}\\ \textbf{Reward} $\uparrow$}
          & \makecell{\textbf{Tetris}\\ \textbf{Reward} $\uparrow$}
          & \makecell{\textbf{CandyCrush}\\ \textbf{Reward} $\uparrow$}
          & \makecell{\textbf{Super Mario Bros}\\ \textbf{Reward} $\uparrow$}
          & \makecell{\textbf{Avg.}\\ \textbf{Reward} $\uparrow$}
          & \makecell{\textbf{Avalon}\\ \textbf{Win Rate} $\uparrow$}
          & \makecell{\textbf{Diplomacy}\\ \textbf{Mean SC} $\uparrow$} \\
          \midrule

          \textsc{GPT-5.4}
          & $\bm{1126.6 \pm 150.2}$
          & $\bm{458.2 \pm 203.5}$
          & $\bm{532.6 \pm 24.8}$
          & $752.0 \pm 35.7$
          & $717.4$
          & $\bm{65.0 \pm 14.2}$
          & $\bm{4.70 \pm 0.35}$ \\

          \textsc{Gemini-3.1-Pro}
          & $813.3 \pm 143.6$
          & $372.7 \pm 157.7$
          & $334.3 \pm 59.4$
          & $436.8 \pm 86.1$
          & $489.3$
          & $42.0 \pm 13.2$
          & $2.72 \pm 0.26$ \\

          \textsc{Claude-4.6-Sonnet}
          & $945.0 \pm 134.5$
          & $444.2 \pm 182.6$
          & $328.6 \pm 23.8$
          & $399.5 \pm 53.4$
          & $529.3$
          & $40.0 \pm 13.1$
          & $3.16 \pm 0.19$ \\

          \textsc{GPT-OSS-120B}
          & $1029.5 \pm 122.0$
          & $358.1 \pm 139.7$
          & $334.4 \pm 40.5$
          & $\bm{968.5 \pm 175.0}$
          & $672.6$
          & $40.0 \pm 13.1$
          & $2.46 \pm 0.25$ \\

        \midrule
          \textsc{Qwen3-8B}
          & $131.0 \pm 102.6$
          & $32.0 \pm 8.5$
          & $519.9 \pm 37.8$
          & $835.5 \pm 161.6$
          & $379.6$
          & $30.0 \pm 9.9$
          & $2.64 \pm 0.18$ \\

        \textsc{\quad SFT w/o Skill}
          & $516.7 \pm 172.3$
          & $28.2 \pm 9.7$
          & $356.1 \pm 30.1$
          & $736.8 \pm 130.4$
          & $409.5$
          & $28.7 \pm 9.7$
          & $2.75 \pm 0.13$ \\

        \textsc{\quad SFT + 1st Skill}
            & $385.5 \pm 239.7$
            & $35.8 \pm 7.0$
            & $569.6 \pm 29.5$
            & $871.9 \pm 126.2$
            & $465.7$
            & $21.2 \pm 8.9$
            & $2.89 \pm 0.20$ \\

        \textsc{\quad SFT + Final Skill}
            & $64.8 \pm 46.0$
            & $24.4 \pm 7.0$
            & $554.4 \pm 24.3$
            & $794.4 \pm 112.9$
            & $359.5$
            & $25.0 \pm 9.3$
            & $2.65 \pm 0.25$ \\

        \textsc{\quad GRPO w/o Skill}
          & $510.0 \pm 249.5$
          & $96.7 \pm 30.3$
          & $163.3 \pm 71.4$
          & $669.4 \pm 130.1$
          & $359.9$
          & $36.2 \pm 10.3$
          & $2.76 \pm 0.19$ \\

        \textsc{\quad GRPO + 1st Skill}
          & $152.0 \pm 107.9$
          & $93.7 \pm 37.7$
          & $353.8 \pm 20.2$
          & $621.2 \pm 130.2$
          & $305.2$
          & $31.2 \pm 10.0$
          & $2.56 \pm 0.22$ \\

        \midrule

          \textsc{\ours{} (Base Qwen3-8B)}
          & $\bm{1589.0 \pm 192.4}$
          & $\bm{510.9 \pm 199.5}$
          & $\bm{648.8 \pm 38.8}$
          & $\bm{948.9 \pm 153.2}$
          & $\bm{924.4}$
          & $\bm{39.0 \pm 9.4}$
          & $\bm{2.96 \pm 0.20}$ \\

          \bottomrule
      \end{tabular}
      }
      \end{center}
      \vspace{-5pt}
        \caption{\textbf{Performance across game categories.}
        We report \textbf{reward} for 2048, Tetris, Candy Crush, and Super Mario Bros, \textbf{overall win rate} for Avalon, and \textbf{overall mean supply centers} for Diplomacy. All results are reported with \textbf{95\% confidence intervals}, based on \textbf{16 evaluation rollouts} for single-player games and \textbf{10 rollouts per player} for multi-player games. \textbf{Conclusion:} \ours{} achieves a significant \textbf{25.1\%} average improvement over GPT-5.4 on single-player games, while remaining competitive with SOTA LLMs on multi-player social reasoning tasks.}
    \label{tab:game_category_results}
    \vspace{-10pt}
\end{table*}

\begin{figure*}[tb]
    \centering
    \vspace{0.1em}
    \includegraphics[width=\textwidth]{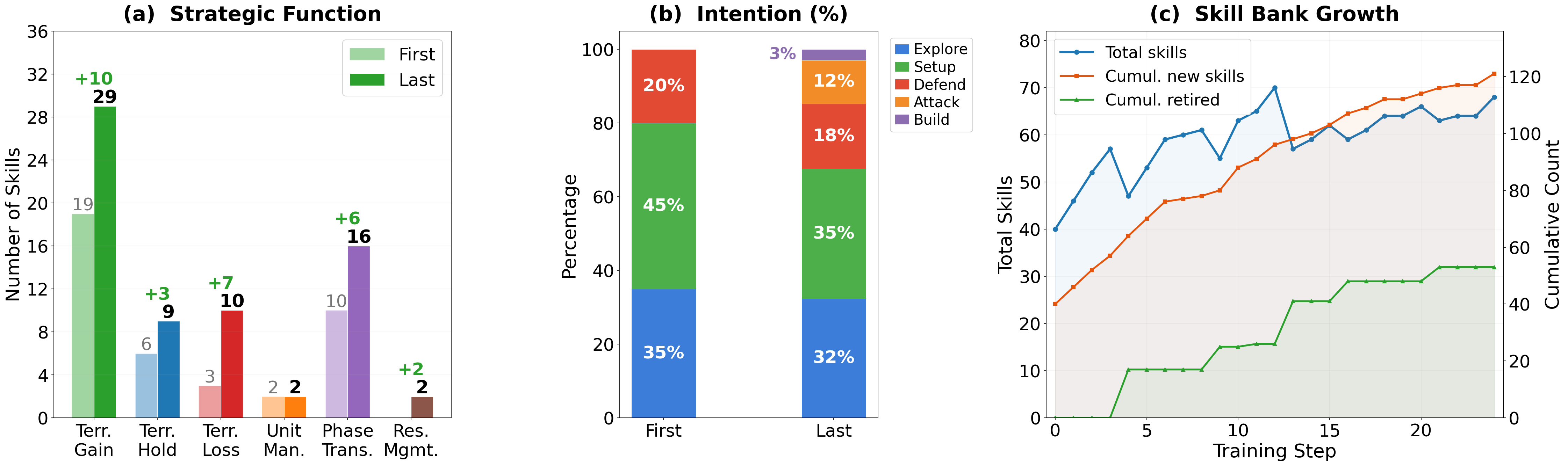}
     \vspace{-15pt}
    \caption{\textbf{Skill bank evolution over Diplomacy training.} \textbf{(a) Development of Strategic Function Categories} from the first to the last training step. Compared with the initial skill bank, the final bank shows notable increases in phase transition and territory loss skills, indicating a broader tactical repertoire. \textbf{(b) Changes in Intention Composition} between the first and last training steps, suggesting increasingly goal-directed behavior and greater diversity across skill categories. \textbf{(c) Skill Bank Growth} during training: the active bank stays at roughly 55--70 skills, while 121 are discovered overall. Periodic curation removes 53 redundant skills via merge and split, keeping the bank compact and effective.}
    \label{fig:diplomacy_skill_bank_evolution}
    \vspace{-12pt}
\end{figure*}

\subsection{Skill Reusability Analysis}

\begin{table*}[t]
    \begin{center}
    \resizebox{0.7\textwidth}{!}{
      \begin{tabular}{lcccccccc}
      \toprule
          \makecell{\textbf{Game}}
          & \makecell{\textbf{\#Skills}}
          & \makecell{\textbf{\#Cats}}
          & \makecell{\textbf{Avg.}\\\textbf{Clauses}}
          & \makecell{\textbf{Tot.}\\\textbf{Sub-Ep}}
          & \makecell{\textbf{Max}\\\textbf{Inst}}
          & \makecell{\textbf{Avg.}\\\textbf{Inst}}
          & \makecell{\textbf{Gini}}
          & \makecell{\textbf{Avg.}\\\textbf{Ver}} \\
          \midrule

          \textsc{2048}
          & 13
          & 6
          & 0.92
          & 593
          & 236
          & 45.6
          & 0.718
          & 3.1 \\

          \textsc{Tetris}
          & 6
          & 2
          & 4.67
          & 295
          & 190
          & 49.2
          & 0.634
          & 4.2 \\

          \textsc{CandyCrush}
          & 6
          & 4
          & 13.67
          & 203
          & 78
          & 33.8
          & 0.498
          & 2.0 \\

          \textsc{Super Mario Bros}
          & 20
          & 9
          & 7.55
          & 281
          & 74
          & 14.1
          & 0.543
          & 2.5 \\

          \textsc{Avalon}
          & 16
          & 3
          & 4.40
          & 479
          & 125
          & 29.9
          & 0.564
          & 6.4 \\

          \textsc{Diplomacy}
          & 64
          & 5
          & 6.52
          & 814
          & 45
          & 12.7
          & 0.524
          & 2.4 \\
          \bottomrule
      \end{tabular}
      }
      \end{center}
      \vspace{-8pt}
    \caption{\textbf{Cross-run skill reusability across games.}
    We report the number of discovered skills (\#Skills), number of categories (\#Cats), average number of clauses per skill (Avg. Clauses), total number of segmented sub-episodes (Tot. Sub-Ep), the maximum number of instances for the most reused skill (Max Inst), the average number of instances per skill (Avg. Inst), reuse concentration (Gini), and the average number of contract refinement versions per skill (Avg. Ver). These results suggest that our framework learns reusable and adaptable skill abstractions rather than memorizing isolated trajectories.}
    \label{tab:skill_reusability_results}
    \vspace{-12pt}
\end{table*}

Apart from decision performance, we also evaluate the \emph{reusability} of the learned skills. Unlike prior approaches that manually specify skills or distill them from strong LLMs with human supervision, our skill-bank agent automatically extracts skills from unlabeled rollout trajectories. A useful skill should both improve downstream rewards and remain reusable across episodes, since repeated reuse suggests stable preconditions, effects, and execution protocols rather than memorization of isolated trajectories.
As summarized in Table~\ref{tab:skill_reusability_results} and illustrated in Figure~\ref{fig:diplomacy_skill_bank_evolution}, our skill bank learns compact and adaptable skill abstractions. In the Diplomacy case study, the bank grows in strategic diversity over training while remaining compact through periodic merge-and-split curation. The x-axis in Figure~\ref{fig:diplomacy_skill_bank_evolution} (left) corresponds to co-evolution iterations, where each point reflects the skill bank after one round of rollout collection and bank update.

\subsection{Discussion}

\textbf{Few-Shot Adaptation with Minimal Loss in General Reasoning.}
Table~\ref{tab:game_category_results} shows that \ours{} substantially improves game-playing performance with only a few adaptation iterations while largely preserving the base model's general reasoning ability. On benchmarks such as Math-500 and MMLU-Pro (Appendix~\ref{app:general_reasoning}), performance drops remain small, suggesting that gains in game environments do not come at the cost of severe reasoning degradation, even though the adapted 8B model reaches strong performance within 25 co-evolution steps.

\textbf{Why Reusable Skills Matter in Long-Horizon Tasks.} Table~\ref{tab:skill_reusability_results} suggests that \ours{} learns reusable skill abstractions rather than memorizing isolated trajectories. This is especially important in long-horizon and socially strategic settings, where coherent multi-step behavior is required. By retrieving only a small set of relevant skills, \ours{} provides compact guidance that improves action selection with low overhead, while continual refinement supports stable co-evolution and efficient adaptation.

\section{Conclusion}
\label{sec:conclusion}


We present \ours{}, a co-evolution framework in which an LLM decision agent retrieves skills from a skill bank to guide action taking, while an agent-managed skill pipeline discovers reusable skills from unlabeled rollouts for future decisions. Our framework improves both sides jointly: the decision agent learns better skill retrieval and action generation, while the skill-bank agent continually extracts, refines, and updates skills and their contracts. By keeping the policy and skill bank aligned over time, \ours{} improves long-horizon control and adaptation. Experiments across diverse game environments show that \ours{} yields strong gains over frontier LLM baselines on single-player games while remaining competitive on multi-player social reasoning tasks.

\textbf{Limitation and Future Work.} A main limitation of \ours{} is its reliance on compact text-based state summaries, which limits grounded multi-modal understanding and can miss critical evidence in raw visual observations or long temporal dependencies. Over long trajectories, summarization errors can accumulate, reducing skill relevance. 
An important direction for future work is to extend \ours{} to multi-modal interactive environments, where learned skills can capture both interaction dynamics and visual reasoning. We also aim to improve cross-domain transfer, so that learned skills generalize across multi-modal games, agentic environments, and broader visual reasoning tasks.

\bibliography{colm2026_conference}
\bibliographystyle{colm2026_conference}

\newpage
\appendix
\section{Reproducibility} 
\label{app:reproduce}
We will release the full \ours{} implementation, including the decision agent, skill bank agent, co-evolution training loop, GRPO training scripts with multi-adapter LoRA configurations, skill bank pipeline, and all prompt templates used in this paper. We will also provide the environment wrappers, reward definitions, and configuration files needed to reproduce the experiments across all six game environments.

We document the full evaluation protocol in the paper and appendix, including environment settings (Appendix~\ref{app:game_envs}), hyperparameters, the number of evaluation rollouts per setting (Appendix~\ref{app:key_hparams}), and reward definitions (Appendix~\ref{app:rwd_design}). For single-player games, we report results over 16 rollouts; for multi-player games, over 10 rollouts per player. Because multi-player experiments use black-box LLM APIs as opponents (GPT-5-mini for training, GPT-5.4 for evaluation), exact numerical replication may vary across API versions and infrastructure conditions. To reduce this variance, we use fixed prompts, deterministic environment seeds where supported, and repeated runs under the same protocol. We therefore expect the main qualitative conclusions to remain reproducible even when exact per-run metrics vary slightly.

\section{Game Environment Settings}
\label{app:game_envs}

We evaluate on six game environments spanning puzzle solving, platform control, and multi-agent social reasoning. In all cases, observations are converted into textual or natural-language state descriptions, and the agent interacts through discrete text actions. Below we summarize the observation space, action space, horizon, and reward for each environment.

\paragraph{2048.}
2048 is a single-player sliding-tile puzzle on a $4\times 4$ grid. Equal-valued tiles merge on contact, and the game ends when no valid moves remain. The agent observes a text-rendered board with tile values, score, maximum tile, empty-cell count, and directional move lookahead. The action space has four actions: \texttt{up}, \texttt{down}, \texttt{left}, and \texttt{right}. Episodes are capped at 200 steps, with reward given by the merge score.

\paragraph{Candy Crush.}
Candy Crush is a single-player match-3 puzzle on an $8\times 8$ board with four candy colors. The player swaps adjacent candies to form matches, after which matched candies are cleared and new candies fall. The agent observes a text-based board and the currently valid swap actions. The action space is dynamic and consists of valid coordinate-pair swaps at each step. Episodes last at most 50 moves, and rewards are based on points from matches and cascades.

\paragraph{Tetris.}
Tetris is a single-player tile-stacking game on a $10\times 20$ board with the standard seven tetrominoes and a four-piece preview queue. The agent receives an ASCII board, the current piece, upcoming pieces, and summary statistics such as stack height, holes, cleared lines, and level. Instead of primitive controls, the environment provides macro actions corresponding to valid placements defined by rotation and target column. Episodes are capped at 200 steps, and reward is the game score.

\paragraph{Super Mario Bros.}
Super Mario Bros. is a single-player side-scrolling platform game where Mario must traverse the level, avoid hazards, and reach the flag. Observations are converted into natural-language descriptions of Mario's position, nearby objects and enemies, and relevant game-state information. The action space contains seven discrete actions: \texttt{noop}, \texttt{right}, \texttt{right+A}, \texttt{A}, \texttt{left}, \texttt{left+A}, and \texttt{down}. Episodes are capped at 200 steps. Rewards combine progress and in-game signals, including distance, coins, and time bonus.

\paragraph{Avalon.}
Avalon is a five-player social deduction game with hidden roles: Merlin, two Servants, one Minion, and one Assassin. Play proceeds through quest rounds with team proposal, voting, quest execution, and assassination phases. The agent observes a natural-language state description containing the phase, private role information, quest progress, leader identity, proposed team, discussion history, and vote outcomes. Actions depend on the phase and include team proposals, votes, quest choices, and assassination targets. Each game contains up to five quests, and rewards are based mainly on whether the agent's side wins.

\paragraph{Diplomacy.}
Diplomacy is a seven-player grand-strategy board game on the standard European map. The powers are Austria, England, France, Germany, Italy, Russia, and Turkey, and the game cycles through movement, retreat, and adjustment phases. The agent observes a natural-language state including the current phase, controlled power, unit locations, supply center counts, valid orders, recent negotiations, and phase history. Actions are standard Diplomacy order strings issued jointly for all controlled units in a phase. Episodes are capped at 20 phases, and rewards are based on the number of controlled supply centers normalized by the victory target of 18.

\section{Key Hyperparameters}
\label{app:key_hparams}

We summarize the main hyperparameters used in co-evolution training for the six game environments in the main paper. Table~\ref{tab:key_hparams} lists the game-specific settings used in our main experiments. All training runs are conducted on an 8$\times$A100 GPU cluster. For games without explicit GRPO overrides, we report the default values directly: GRPO clip ratio $0.2$, maximum 4 epochs, no advantage clipping, learning rate $5\times 10^{-5}$, and KL coefficient $0.05$.

\begin{table*}[h]
    \vspace{0.4em}
    \begin{center}
    \resizebox{\textwidth}{!}{
    \begin{tabular}{lcccccccc}
    \toprule
    \makecell{\textbf{Game}}
    & \makecell{\textbf{Total}\\\textbf{Steps}}
    & \makecell{\textbf{Episodes}\\\textbf{/ Step}}
    & \makecell{\textbf{Ckpt}\\\textbf{Int.}}
    & \makecell{\textbf{GRPO}\\\textbf{LR}}
    & \makecell{\textbf{KL}\\\textbf{Coeff.}}
    & \makecell{\textbf{Clip}}
    & \makecell{\textbf{Max}\\\textbf{Epochs}}
    & \makecell{\textbf{Adv.}\\\textbf{Clip}} \\
    \midrule
    2048
    & 10 & 8 & 3
    & $5\!\times\!10^{-5}$
    & 0.05
    & 0.20
    & 4
    & -- \\

    Candy Crush
    & 10 & 8 & 3
    & $5\!\times\!10^{-5}$
    & 0.05
    & 0.20
    & 4
    & -- \\

    Tetris
    & 7 & 8 & 1
    & $2\!\times\!10^{-5}$
    & 0.08
    & 0.10
    & 2
    & 3.0 \\

    Super Mario Bros.
    & 20 & 8 & 1
    & $3\!\times\!10^{-5}$
    & 0.04
    & 0.15
    & 3
    & 5.0 \\

    Avalon
    & 20 & 20 & 1
    & $2\!\times\!10^{-5}$
    & 0.06
    & 0.15
    & 2
    & 3.0 \\

    Diplomacy
    & 25 & 28 & 1
    & $1\!\times\!10^{-5}$
    & 0.08
    & 0.12
    & 2
    & 3.0 \\
    \bottomrule
    \end{tabular}}
    \end{center}
    \caption{Key co-evolution training hyperparameters across the six game environments in the main paper. ``Ckpt Int.'' denotes checkpoint interval. For 2048 and Candy Crush, all GRPO fields use the default values.}
    \label{tab:key_hparams}
    \vspace{-0.8em}
\end{table*}

\section{Breakdown of Role-wise Performance on Multi-player Games}
\label{app:more_results}

For multi-player social games, we use \textsc{GPT-5.4} as the opponent model. We report the main results in Table~\ref{tab:game_category_results}, with role-wise and power-wise breakdowns provided in Table~\ref{tab:avalon_per_role_winrate_compare} and Table~\ref{tab:diplomacy_per_power_mean_sc_ci}.

\begin{table*}[h]
    \vspace{0.6em}
    \begin{center}
    \resizebox{\textwidth}{!}{
      \begin{tabular}{lccc|ccc|c}
      \toprule
          \multicolumn{1}{c}{}
          & \multicolumn{6}{c|}{\makecell{\textbf{Avalon}}}
          & \multicolumn{1}{c}{\makecell{\textbf{Overall}}} \\
          \cmidrule(lr){2-7}\cmidrule(lr){8-8}
          \makecell{\textbf{Model}}
          & \multicolumn{3}{c|}{\makecell{\textbf{Good}}}
          & \multicolumn{3}{c|}{\makecell{\textbf{Evil}}}
          & \makecell{\textbf{Avg.}\\ \textbf{Win Rate} $\uparrow$} \\
          \cmidrule(lr){2-4}\cmidrule(lr){5-7}
          & \makecell{\textbf{Merlin}\\ \textbf{Win Rate} $\uparrow$}
          & \makecell{\textbf{Servant}\\ \textbf{Win Rate} $\uparrow$}
          & \makecell{\textbf{Good Avg.}\\ \textbf{Win Rate} $\uparrow$}
          & \makecell{\textbf{Assassin}\\ \textbf{Win Rate} $\uparrow$}
          & \makecell{\textbf{Minion}\\ \textbf{Win Rate} $\uparrow$}
          & \makecell{\textbf{Evil Avg.}\\ \textbf{Win Rate} $\uparrow$}
          & \\
          \midrule

          \textsc{GPT-5.4}
          & $\bm{62.5 \pm 27.9}$
          & $\bm{47.4 \pm 20.5}$
          & $\bm{51.9 \pm 17.7}$
          & $\bm{100.0 \pm 19.5}$
          & $\bm{85.7 \pm 24.4}$
          & $\bm{92.3 \pm 15.9}$
          & $\bm{65.0 \pm 14.2}$ \\

          \textsc{Gemini-3.1-Pro}
          & $10.0 \pm 19.3$
          & $36.4 \pm 18.7$
          & $28.1 \pm 14.9$
          & $66.7 \pm 30.2$
          & $66.7 \pm 23.6$
          & $66.7 \pm 20.0$
          & $42.0 \pm 13.2$ \\

          \textsc{Claude-4.6}
          & $30.0 \pm 24.8$
          & $40.9 \pm 19.0$
          & $37.5 \pm 15.9$
          & $33.3 \pm 30.2$
          & $50.0 \pm 24.6$
          & $44.4 \pm 20.9$
          & $40.0 \pm 13.1$ \\

          \textsc{GPT-OSS-120B}
          & $30.0 \pm 24.8$
          & $31.8 \pm 18.2$
          & $31.2 \pm 15.3$
          & $50.0 \pm 31.2$
          & $58.3 \pm 24.4$
          & $55.6 \pm 20.9$
          & $40.0 \pm 13.1$ \\

          \midrule

          \textsc{Qwen3-8B}
          & $18.8 \pm 18.2$
          & $18.4 \pm 12.1$
          & $18.5 \pm 10.2$
          & $\bm{66.7 \pm 23.6}$
          & $42.9 \pm 23.0$
          & $53.8 \pm 17.9$
          & $30.0 \pm 9.9$ \\

          \textsc{\quad SFT w/o Skill}
          & $18.8 \pm 18.2$
          & $18.4 \pm 12.1$
          & $18.5 \pm 10.2$
          & $41.7 \pm 24.4$
          & $57.1 \pm 23.0$
          & $50.0 \pm 17.9$
          & $28.7 \pm 9.7$ \\

          \textsc{\quad SFT + 1st Skill}
          & $6.2 \pm 13.6$
          & $13.2 \pm 10.8$
          & $11.1 \pm 8.5$
          & $33.3 \pm 23.6$
          & $50.0 \pm 23.2$
          & $42.3 \pm 17.8$
          & $21.2 \pm 8.9$ \\

          \textsc{\quad SFT + Final Skill}
          & $25.0 \pm 19.7$
          & $\bm{26.3 \pm 13.5}$
          & $25.9 \pm 11.4$
          & $8.3 \pm 17.0$
          & $35.7 \pm 22.4$
          & $23.1 \pm 15.5$
          & $25.0 \pm 9.3$ \\

          \textsc{\quad GRPO w/o Skill}
          & $\bm{37.5 \pm 21.4}$
          & $15.8 \pm 11.5$
          & $22.2 \pm 10.9$
          & $58.3 \pm 24.4$
          & $\bm{71.4 \pm 21.5}$
          & $\bm{65.4 \pm 17.2}$
          & $36.2 \pm 10.3$ \\

          \textsc{\quad GRPO + 1st Skill}
          & $31.2 \pm 20.7$
          & $18.4 \pm 12.1$
          & $22.2 \pm 10.9$
          & $41.7 \pm 24.4$
          & $57.1 \pm 23.0$
          & $50.0 \pm 17.9$
          & $31.2 \pm 10.0$ \\

        \midrule
          \textsc{\ours{} (Base Qwen3-8B)}
          & $33.3 \pm 17.7$
          & $23.3 \pm 12.2$
          & $\bm{26.9 \pm 10.4}$
          & $64.3 \pm 22.5$
          & $63.2 \pm 20.0$
          & $63.6 \pm 15.6$
          & $\bm{39.0 \pm 9.4}$ \\

          \bottomrule
      \end{tabular}
      }
      \end{center}
      \vspace{-5pt}
    \caption{\textbf{Per-role win rates for Avalon.}
    We report win rates with 95\% confidence for both good-side roles (Merlin, Servant), evil-side roles (Assassin, Minion). Results show that our method reaches comparable performance to \textsc{Gemini-3.1-Pro} and \textsc{GPT-OSS-120B}.}
    \label{tab:avalon_per_role_winrate_compare}
    \vspace{-10pt}
\end{table*}
\begin{table*}[h]
    \vspace{0.6em}
    \begin{center}
    \resizebox{\textwidth}{!}{
      \begin{tabular}{lccccccc|c}
      \toprule
          \multicolumn{1}{c}{}
          & \multicolumn{7}{c}{\makecell{\textbf{Diplomacy}}}
          & \multicolumn{1}{c}{\makecell{\textbf{Overall}}} \\
          \cmidrule(lr){2-8}\cmidrule(lr){9-9}
          \makecell{\textbf{Model}}
          & \makecell{\textbf{Austria}\\ \textbf{Mean SC} $\uparrow$}
          & \makecell{\textbf{England}\\ \textbf{Mean SC} $\uparrow$}
          & \makecell{\textbf{France}\\ \textbf{Mean SC} $\uparrow$}
          & \makecell{\textbf{Germany}\\ \textbf{Mean SC} $\uparrow$}
          & \makecell{\textbf{Italy}\\ \textbf{Mean SC} $\uparrow$}
          & \makecell{\textbf{Russia}\\ \textbf{Mean SC} $\uparrow$}
          & \makecell{\textbf{Turkey}\\ \textbf{Mean SC} $\uparrow$}
          & \makecell{\textbf{Mean}\\ \textbf{SC} $\uparrow$} \\
          \midrule

          \textsc{GPT-5.4}
          & $\bm{4.38 \pm 1.34}$
          & $\bm{4.12 \pm 0.82}$
          & $\bm{4.50 \pm 1.09}$
          & $\bm{5.12 \pm 0.95}$
          & $\bm{4.50 \pm 0.77}$
          & $\bm{5.12 \pm 1.52}$
          & $\bm{5.12 \pm 0.95}$
          & $\bm{4.70 \pm 0.35}$ \\

        \textsc{Gemini-3.1-Pro}
          & $2.50 \pm 0.89$
          & $2.38 \pm 0.63$
          & $3.12 \pm 0.70$
          & $2.88 \pm 0.82$
          & $3.14 \pm 0.35$
          & $3.14 \pm 0.35$
          & $1.86 \pm 1.24$
          & $2.72 \pm 0.26$ \\

          \textsc{Claude-4.6}
          & $3.20 \pm 0.56$
          & $2.90 \pm 0.41$
          & $3.50 \pm 0.70$
          & $3.78 \pm 0.64$
          & $3.20 \pm 0.30$
          & $2.80 \pm 0.30$
          & $2.80 \pm 0.66$
          & $3.16 \pm 0.19$ \\

          \textsc{GPT-OSS-120B}
          & $1.75 \pm 1.07$
          & $2.88 \pm 0.29$
          & $2.88 \pm 0.29$
          & $2.62 \pm 0.44$
          & $3.00 \pm 0.00$
          & $2.88 \pm 0.29$
          & $1.25 \pm 0.97$
          & $2.46 \pm 0.25$ \\

        \midrule

        \textsc{Qwen3-8B}
          & $1.62 \pm 0.26$
          & $2.75 \pm 0.25$
          & $2.75 \pm 0.16$
          & $2.88 \pm 0.12$
          & $3.00 \pm 0.00$
          & $2.88 \pm 0.23$
          & $2.62 \pm 0.18$
          & $2.64 \pm 0.18$ \\

          \textsc{\quad SFT w/o skill}
          & $2.12 \pm 0.23$
          & $2.75 \pm 0.16$
          & $2.88 \pm 0.12$
          & $2.62 \pm 0.26$
          & $2.88 \pm 0.12$
          & $3.00 \pm 0.00$
          & $3.00 \pm 0.00$
          & $2.75 \pm 0.13$ \\

          \textsc{\quad SFT + 1st skill}
          & $2.12 \pm 0.30$
          & $2.88 \pm 0.12$
          & $\bm{3.12 \pm 0.30}$
          & $2.50 \pm 0.33$
          & $3.12 \pm 0.12$
          & $3.25 \pm 0.25$
          & $\bm{3.25 \pm 0.16}$
          & $2.89 \pm 0.20$ \\

          \textsc{\quad SFT + final skill}
          & $2.12 \pm 0.44$
          & $2.75 \pm 0.16$
          & $2.25 \pm 0.25$
          & $2.71 \pm 0.29$
          & $2.88 \pm 0.23$
          & $2.88 \pm 0.12$
          & $3.00 \pm 0.57$
          & $2.65 \pm 0.25$ \\

          \textsc{\quad GRPO w/o skill}
          & $\bm{2.33 \pm 0.33}$
          & $2.83 \pm 0.17$
          & $3.00 \pm 0.00$
          & $2.80 \pm 0.20$
          & $3.00 \pm 0.00$
          & $3.20 \pm 0.20$
          & $2.20 \pm 0.37$
          & $2.76 \pm 0.19$ \\

          \textsc{\quad GRPO + 1st skill}
          & $1.38 \pm 0.18$
          & $\bm{3.12 \pm 0.12}$
          & $2.25 \pm 0.16$
          & $2.75 \pm 0.37$
          & $3.00 \pm 0.00$
          & $3.25 \pm 0.16$
          & $2.14 \pm 0.26$
          & $2.56 \pm 0.22$ \\

        \midrule
          \textsc{\ours{} (Base Qwen3-8B)}
          & $2.22 \pm 0.55$
            & $2.70 \pm 0.42$
            & $3.10 \pm 0.46$
            & $\bm{3.12 \pm 0.58}$
            & $\bm{3.20 \pm 0.39}$
            & $\bm{3.30 \pm 0.30}$
            & $3.00 \pm 0.72$
            & $\bm{2.96 \pm 0.20}$ \\

          \bottomrule
      \end{tabular}
      }
      \end{center}
      \vspace{-5pt}
    \caption{\textbf{Per-power mean supply centers (SC) for Diplomacy.}
    We report the mean supply centers achieved by each model for each power, together with the overall mean across all powers, with 95\% confidence intervals. GPT-5.4 is evaluated in self-play, while all other models are evaluated against GPT-5.4. \textbf{Conclusion:} \ours{} outperforms \textsc{Gemini-3.1-Pro} by 8.8\% on average.}
    \label{tab:diplomacy_per_power_mean_sc_ci}
    \vspace{-10pt}
\end{table*}

Avalon is a team-based competitive game in which only one side can win. It is structurally harder for the Good side, since Good players must infer hidden roles from sparse signals such as proposals, votes, and quest outcomes, while Evil players begin with full coordination and can strategically hide or sabotage~\citep{light2023avalonbench}. As shown in Table~\ref{tab:game_category_results}, our method achieves performance comparable to \textsc{Gemini-3.1-Pro} and \textsc{gpt-oss-120b}, with a slight 1\% lower in win rate. This setting highlights a key weakness of baseline LLMs: they often struggle to maintain consistent beliefs, track intent across rounds, and act coherently from prior evidence. Our method alleviates this by converting raw histories into structured states and applying phase-specific skills, enabling more consistent suspicion modeling and multi-round commitment.

Diplomacy further amplifies these challenges, as it requires long-context reasoning, negotiation, alliance tracking, and multi-phase planning. While baselines often fall back to defensive, non-cooperative behavior against strong opponents, our method benefits from iterative skill-bank refinement, which provides reusable strategic patterns and improves few-shot adaptation over repeated interactions, inducing at least 8.8\% better performance over \textsc{Gemini-3.1-Pro}. Overall, social games expose the weakness of stateless LLM agents, whereas our framework supports long-horizon social reasoning through persistent state tracking, cross-turn memory, and skill-guided decision-making.

\section{Necessity of Using Different LoRA}
\label{app:3-in-1}
\begin{table}[h]
    \centering
    \resizebox{0.5\linewidth}{!}{
    \begin{tabular}{lc}
    \toprule
    \makecell{\textbf{Model}} & \makecell{\textbf{Reward} $\uparrow$} \\
    \midrule
    \textsc{GPT-5.4} & $532.6 \pm 24.8$ \\
    \textsc{Qwen3-8B} & $519.9 \pm 37.8$ \\
    \textsc{\ours{} (Merged LoRAs)} & $502.5 \pm 22.4$ \\
    \textsc{\ours{} (Base Qwen3-8B)} & $648.8 \pm 38.8$ \\
    \bottomrule
    \end{tabular}
    }
    \caption{\textbf{The necessity of splitting LoRAs.} We conduct a supplementary experiment on \textsc{Candy Crush}, comparing \textsc{\ours{}} with stage-specific LoRAs against a merged variant that uses one LoRA for the decision agent and one for the skill-bank agent. The merged-LoRA variant performs worse during training, indicating that splitting LoRAs is important for handling the diverse objectives arising in co-evolution.}
    \label{tab:candycrush_subtable}
\end{table}
In our framework, both the decision agent and the skill-bank agent use multiple LoRA adapters for different functional stages. Specifically, the decision agent uses two LoRAs for skill retrieval and action generation, while the skill-bank agent uses three LoRAs for trajectory segmentation, contract generation, and skill-bank maintenance. To better understand the role of each adapter and the effect of splitting LoRAs by function, we conduct a supplementary experiment on \textsc{Candy Crush} in Table~\ref{tab:candycrush_subtable}, comparing \textsc{\ours{}} with stage-specific LoRAs against a merged variant that uses only one LoRA for the decision agent and one for the skill-bank agent.
The results show that the merged-LoRA variant performs worse during training, suggesting that splitting LoRAs is important for co-evolution. When diverse objectives are compressed into a single adapter, the learned capacity must be shared across incompatible functions, which can cause interference. As shown in Table~\ref{tab:candycrush_subtable}, stage-specific LoRAs enable cleaner specialization, reduce optimization conflict, and better preserve functional capabilities over time.

\section{Reward Design}
\label{app:rwd_design}

Both the decision agent and the skill bank agent are trained with GRPO using function-specific LoRA adapters. Each adapter has a dedicated reward function designed to reinforce the specific capability it governs. All reward computations are CPU-only and require no additional LLM inference. 


\subsection{Decision Agent Rewards}




\paragraph{Action-Taking.}
The action-taking adapter is trained with a per-step reward
\begin{align}
    r_t = r_t^{\mathrm{env}} + \lambda_{\mathrm{f}}\, r_t^{\mathrm{follow}} + r_t^{\mathrm{cost}},
    \label{eq:action_reward}
\end{align}
where $r_t^{\mathrm{env}}$ is the environment reward, $r_t^{\mathrm{follow}}$ is a skill-following shaping term, and $r_t^{\mathrm{cost}}$ penalizes skill switching. 
The shaping term densifies sparse rewards during skill execution: satisfying a new predicate in the active skill’s effect contract gives a small bonus, satisfying all predicates gives a larger bonus, and steps with no progress receive a small penalty. 
The cost term applies a negative reward to skill switching, which is a common failure mode in early training. We set $\lambda_f = 0.1$ in all experiments.

\paragraph{Skill Retrieval.}
The skill-retrieval adapter is trained with a delayed reward evaluated at skill-switch time.
The reward is computed from the normalized environment reward accumulated during the skill, a temporal efficiency term, a contract-completion term, an abort penalty for violated preconditions, and a low-weight retrieval-confidence prior from the RAG module.
This reward favors skills that are successful, efficient, and applicable in the current state.

\subsection{Skill Bank Agent Rewards}



\paragraph{Infer Segmentation.}
The segmentation adapter is trained to assign high-reward trajectory segments to reusable skills in the current bank.
Its reward includes: (i) the fraction of positive episode reward covered by existing skills, (ii) a value-matching term between skill quality and segment importance, (iii) the Viterbi decode score for global consistency, and (iv) a decode-margin term for confident assignments.
Assignments to \emph{new skill} receive partial credit, preventing systematic overuse of existing skills when new behavior appears.

\paragraph{Contract Learning.}
The contract adapter is trained to produce compact effect contracts for each skill.
Without held-out instances, the reward uses F1 between predicted effects and observed start/end predicate changes, with additional terms for added/deleted predicate coverage, sparsity, and specificity.
With held-out instances, the reward emphasizes generalization through a reward-weighted holdout pass rate, together with a precision/recall term against the consensus effects and a reward-alignment term favoring contracts that hold on high-value instances.



\paragraph{Skill Bank Curator.}
The curator adapter evaluates proposed bank updates, including materialize, merge, split, and retire operations.
The reward includes: (i) a quality-alignment term from the continuous skill-quality score, (ii) an exploration term for approving promising new skills with adequate evidence, and (iii) a reason-quality term that favors evidence-based decisions using pass rates, quality scores, and instance counts.
The exploration term helps prevent systematic rejection of new skills.



\begin{wraptable}{r}{0.48\textwidth}
    \vspace{-0.6em}
    \centering
    \small
    \begin{tabular}{lcc}
    \toprule
        \textbf{Model}
        & \makecell{\textbf{MMLU-Pro} \\ \textbf{Acc.} $\uparrow$}
        & \makecell{\textbf{Math-500} \\ \textbf{EM} $\uparrow$} \\
        \midrule
        \textsc{Qwen3-8B} & 61.99\% & 46.40\% \\
        \textsc{\ours{}}     & 61.15\%          & 44.60\%          \\
        \bottomrule
    \end{tabular}
    \vspace{-4pt}
    \caption{\textbf{Overall performance on general reasoning benchmarks.}
    \textsc{\ours{}} remains comparable to the base model on both benchmarks.}
    \label{tab:general_reasoning_summary}
\end{wraptable}

\section{Generalization onto General LLM Reasoning Tasks}
\label{app:general_reasoning}
We further evaluate \textsc{\ours{}} on two standard LLM reasoning benchmarks, Math-500~\cite{hendrycks2021math, lightman2023letsverify} and MMLU-Pro~\cite{wang2024mmlu}, to assess whether our adaptation preserves the base model's general reasoning ability. Math-500 measures mathematical reasoning, while MMLU-Pro evaluates broad knowledge and reasoning across diverse domains. As shown in Table~\ref{tab:general_reasoning_summary}, \textsc{\ours{}} achieves performance close to \textsc{Qwen3-8B} on both benchmarks, with absolute drops of only 0.8\% on MMLU-Pro and 1.8\% on Math-500, indicating that our method largely preserves general reasoning performance.

\section{Training Curve}
\label{app:training_curve}
\begin{figure*}[h]
    \centering
    \vspace{0.1em}
    \includegraphics[width=\textwidth]{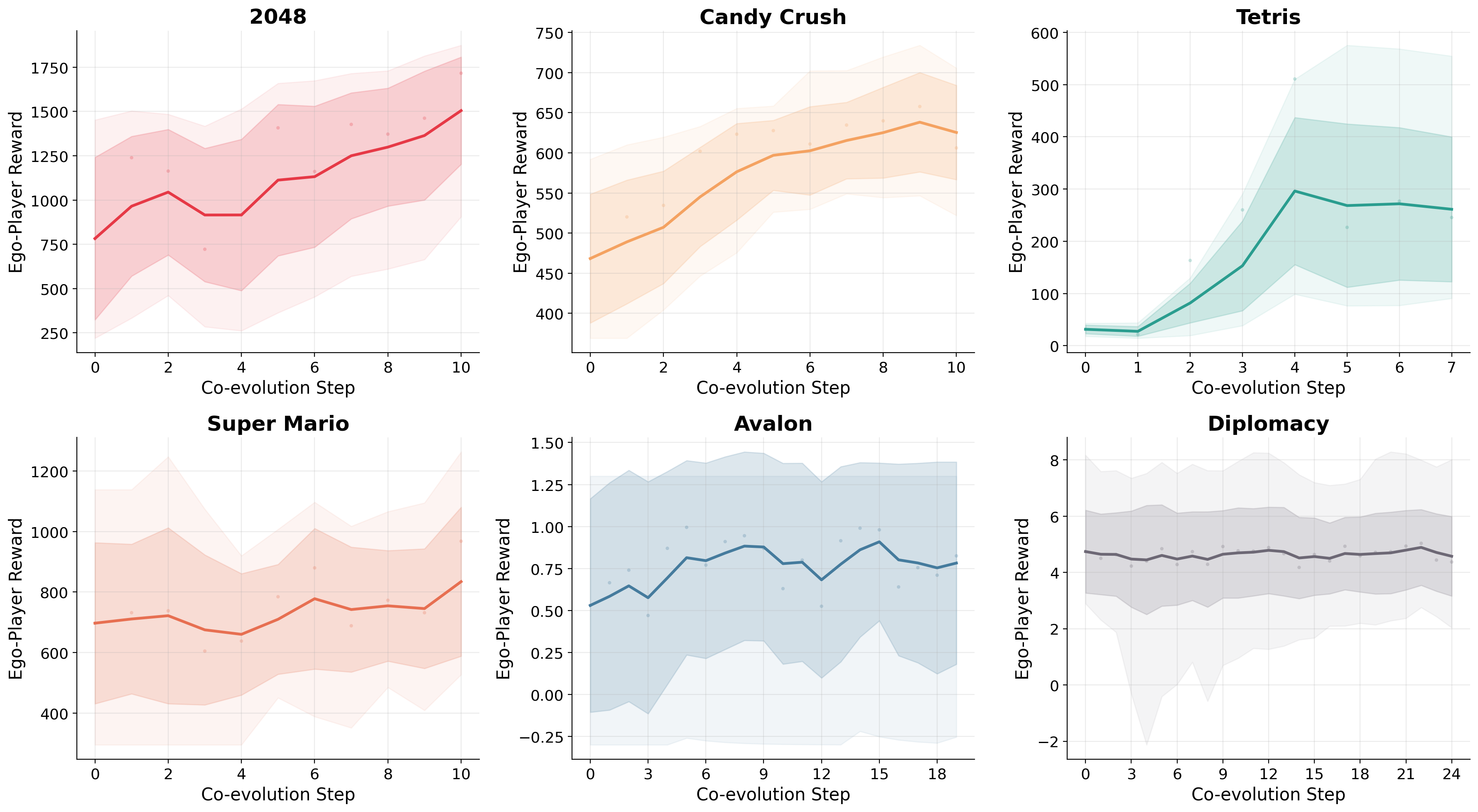}
     \vspace{-7pt}
    \caption{\textbf{Co-evolution reward curves for all games.} Single-player games show steady gains, indicating improved strategies from joint decision-agent and skill-bank training. Multiplayer self-play remains flat because all players improve symmetrically, pushing rewards toward equilibrium.}
    \label{fig:training}
    \vspace{-12pt}
\end{figure*}

Figure~\ref{fig:training} presents the training reward curves for all six game environments. In the single-player games (2048, Candy Crush, Tetris, and Super Mario), the ego-player reward increases steadily over co-evolution steps, confirming that the joint optimization of the decision agent and skill bank yields progressively stronger play. The multi-player games (Avalon and Diplomacy), trained under self-play, exhibit flatter reward trajectories. This is expected: since all players share the same model, improvements to the ego agent are mirrored by equally stronger opponents, causing the reward to converge near the game-theoretic equilibrium rather than rise monotonically. Notably, the variance bands in Diplomacy remain tight throughout training, suggesting stable learning dynamics despite the complex 7-player negotiation setting. The true policy improvement in these competitive games is revealed not by the self-play reward curve but by evaluation against external opponents, as reported in Table.~\ref{tab:game_category_results}.

\section{Qualitative Analysis}
\label{app:qualitative_analysis}
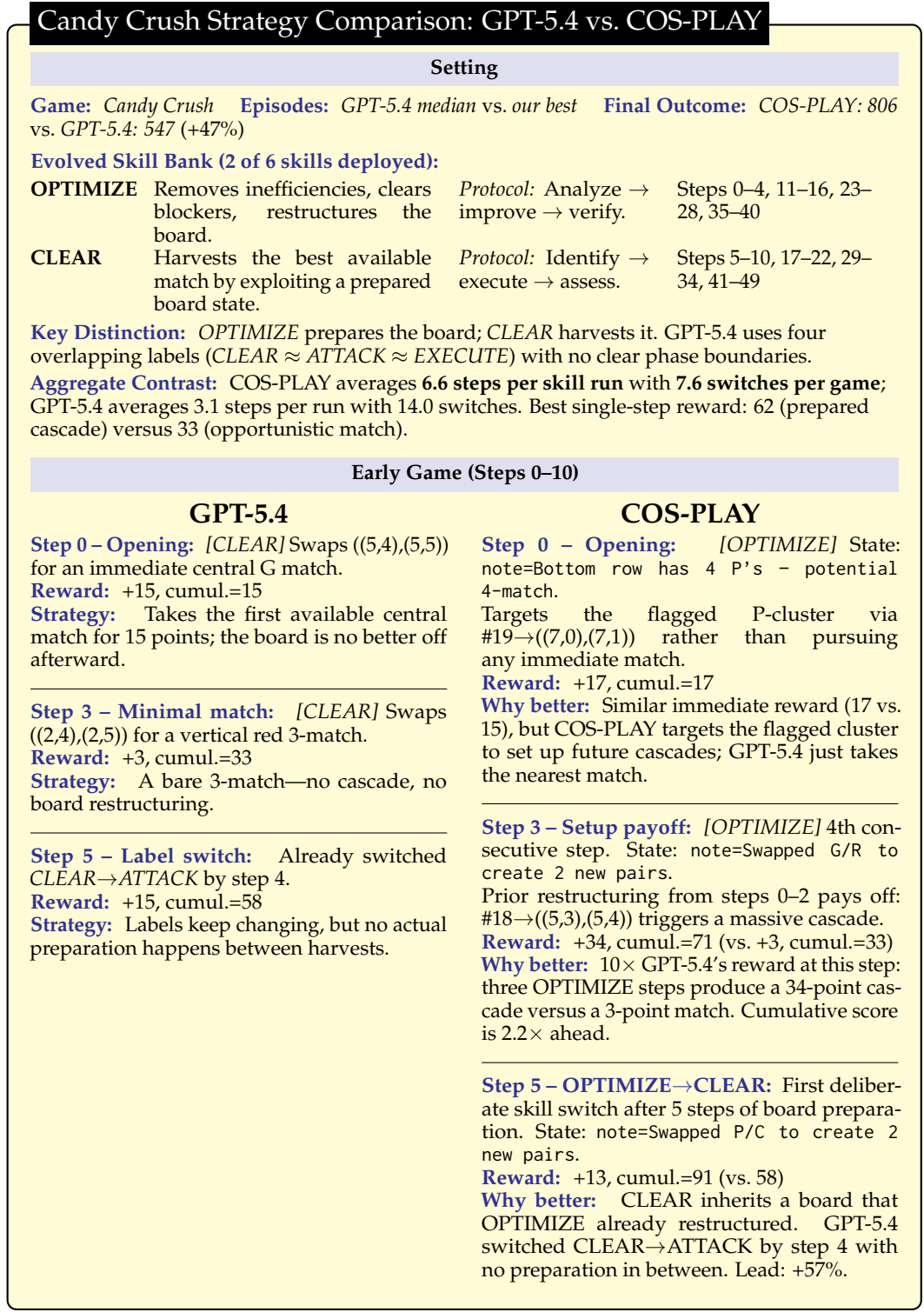
\begin{figure}[!p]
    \begin{center}
    \begin{tikzpicture}
    \node [mybox,title=Candy Crush Strategy Comparison: GPT-5.4 vs.\ \ours{}] (box){%
        \begin{minipage}{0.95\textwidth}
            \footnotesize
            \raggedright

            \noindent\colorbox{Blue!10}{\parbox{\dimexpr\linewidth-2\fboxsep}{\centering\small\textbf{Setting}}}
            \\[4pt]

            \smallbluebold{Game: } \textit{Candy Crush} \quad
            \smallbluebold{Episodes: } \textit{GPT-5.4 median} vs.\ \textit{our best} \quad
            \smallbluebold{Final Outcome: } \textit{\ours{}: 806} vs.\ \textit{GPT-5.4: 547} (+47\%)
            \\[4pt]

        \smallbluebold{Evolved Skill Bank (2 of 6 skills deployed):} \\[2pt]
        \begin{tabular}{@{}l@{\hspace{0.8em}}p{0.32\textwidth}p{0.22\textwidth}p{0.22\textwidth}@{}}
        \textbf{OPTIMIZE} & Removes inefficiencies, clears blockers, restructures the board. &
        \textit{Protocol:} Analyze $\rightarrow$ improve $\rightarrow$ verify. &
        Steps 0--4, 11--16, 23--28, 35--40 \\
        \textbf{CLEAR} & Harvests the best available match by exploiting a prepared board state. &
        \textit{Protocol:} Identify $\rightarrow$ execute $\rightarrow$ assess. &
        Steps 5--10, 17--22, 29--34, 41--49
        \end{tabular} \\[2pt]
        \smallbluebold{Key Distinction: } \textit{OPTIMIZE} prepares the board; \textit{CLEAR} harvests it.
        GPT-5.4 uses four overlapping labels (\textit{CLEAR $\approx$ ATTACK $\approx$ EXECUTE}) with no clear phase boundaries.
        \\[2pt]
        \smallbluebold{Aggregate Contrast: }
        \ours{} averages \textbf{6.6 steps per skill run} with \textbf{7.6 switches per game};
        GPT-5.4 averages 3.1 steps per run with 14.0 switches.
        Best single-step reward: 62 (prepared cascade) versus 33 (opportunistic match).
        \\[6pt]

            \noindent\colorbox{Blue!10}{\parbox{\dimexpr\linewidth-2\fboxsep}{\centering\small\textbf{Early Game (Steps 0--10)}}}
            \\[4pt]

            \noindent
            \begin{minipage}[t]{0.48\linewidth}
            {\centering\textbf{\large GPT-5.4}\par}\vspace{0.3em}
            \smallbluebold{Step 0 -- Opening: }
            \textit{[CLEAR]} Swaps ((5,4),(5,5)) for an immediate central G match. \\
            \smallbluebold{Reward: } +15, cumul.=15 \\
            \smallbluebold{Strategy: } Takes the first available central match for 15 points; the board is no better off afterward.

            \noindent\rule{\linewidth}{0.4pt}\vspace{0.3em}

            \smallbluebold{Step 3 -- Minimal match: }
            \textit{[CLEAR]} Swaps ((2,4),(2,5)) for a vertical red 3-match. \\
            \smallbluebold{Reward: } +3, cumul.=33 \\
            \smallbluebold{Strategy: } A bare 3-match---no cascade, no board restructuring.

            \noindent\rule{\linewidth}{0.4pt}\vspace{0.3em}

            \smallbluebold{Step 5 -- Label switch: }
            Already switched \textit{CLEAR}$\rightarrow$\textit{ATTACK} by step~4. \\
            \smallbluebold{Reward: } +15, cumul.=58 \\
            \smallbluebold{Strategy: } Labels keep changing, but no actual preparation happens between harvests.
            \end{minipage}\hfill
            \begin{minipage}[t]{0.48\linewidth}
            {\centering\textbf{\large \ours{}}\par}\vspace{0.3em}
            \smallbluebold{Step 0 -- Opening: }
            \textit{[OPTIMIZE]} State: \texttt{note=Bottom row has 4~P's -- potential 4-match}. \\
            Targets the flagged P-cluster via \#19$\rightarrow$((7,0),(7,1)) rather than pursuing any immediate match. \\
            \smallbluebold{Reward: } +17, cumul.=17 \\
            \smallbluebold{Why better: } Similar immediate reward (17 vs.\ 15), but \ours{} targets the flagged cluster to set up future cascades; GPT-5.4 just takes the nearest match.

            \noindent\rule{\linewidth}{0.4pt}\vspace{0.3em}

            \smallbluebold{Step 3 -- Setup payoff: }
            \textit{[OPTIMIZE]} 4th consecutive step. State: \texttt{note=Swapped G/R to create 2 new pairs}. \\
            Prior restructuring from steps 0--2 pays off: \#18$\rightarrow$((5,3),(5,4)) triggers a massive cascade. \\
            \smallbluebold{Reward: } +34, cumul.=71 (vs.\ +3, cumul.=33) \\
            \smallbluebold{Why better: } 10$\times$ GPT-5.4's reward at this step: three OPTIMIZE steps produce a 34-point cascade versus a 3-point match. Cumulative score is 2.2$\times$ ahead.

            \noindent\rule{\linewidth}{0.4pt}\vspace{0.3em}

            \smallbluebold{Step 5 -- OPTIMIZE$\rightarrow$CLEAR: }
            First deliberate skill switch after 5 steps of board preparation. State: \texttt{note=Swapped P/C to create 2 new pairs}. \\
            \smallbluebold{Reward: } +13, cumul.=91 (vs.\ 58) \\
            \smallbluebold{Why better: } CLEAR inherits a board that OPTIMIZE already restructured. GPT-5.4 switched CLEAR$\to$ATTACK by step~4 with no preparation in between. Lead: +57\%.
            \end{minipage}

        \end{minipage}
    };
    \end{tikzpicture}%
    \caption{\textbf{Step-level comparison between GPT-5.4 and our method in Candy Crush} (1/2: Setting \& Early Game).
    Continued in Figure~\ref{candy-crush-comparison-cont}.}
    \label{candy-crush-comparison}
    \end{center}
\end{figure}

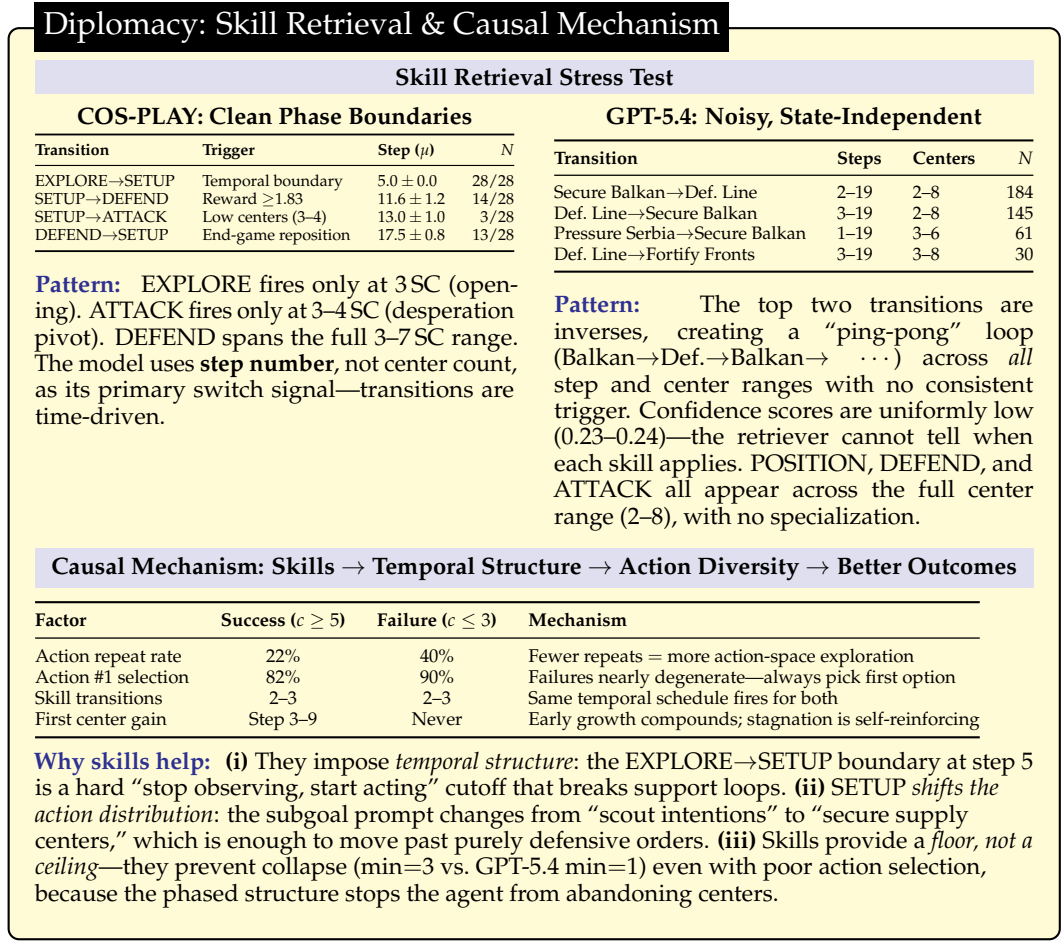
\begin{figure*}[!p]
    \begin{center}
    \begin{tikzpicture}
    \node [mybox,title={Diplomacy: Skill Retrieval \& Causal Mechanism}] (box){%
        \begin{minipage}{0.95\textwidth}
            \footnotesize
            \raggedright

            \noindent\colorbox{Blue!10}{\parbox{\dimexpr\linewidth-2\fboxsep}{\centering\small\textbf{Skill Retrieval Stress Test}}}
            \\[4pt]

            \noindent
            \begin{minipage}[t]{0.48\linewidth}
            {\centering\small\textbf{\ours{}: Clean Phase Boundaries}\par}\vspace{0.3em}

            \resizebox{\linewidth}{!}{\scriptsize
            \begin{tabular}{@{}lllr@{}}
            \toprule
            \textbf{Transition} & \textbf{Trigger} & \textbf{Step ($\mu$)} & \textbf{$N$} \\
            \midrule
            EXPLORE$\to$SETUP & Temporal boundary & $5.0 \pm 0.0$ & 28/28 \\
            SETUP$\to$DEFEND & Reward ${\geq}1.83$ & $11.6 \pm 1.2$ & 14/28 \\
            SETUP$\to$ATTACK & Low centers (3--4) & $13.0 \pm 1.0$ & 3/28 \\
            DEFEND$\to$SETUP & End-game reposition & $17.5 \pm 0.8$ & 13/28 \\
            \bottomrule
            \end{tabular}}
            \\[4pt]

            \smallbluebold{Pattern: } EXPLORE fires only at 3\,SC (opening).
            ATTACK fires only at 3--4\,SC (desperation pivot).
            DEFEND spans the full 3--7\,SC range.
            The model uses \textbf{step number}, not center count, as its primary switch signal---transitions are time-driven.
            \end{minipage}\hfill
            \begin{minipage}[t]{0.48\linewidth}
            {\centering\small\textbf{GPT-5.4: Noisy, State-Independent}\par}\vspace{0.3em}

            \resizebox{\linewidth}{!}{\scriptsize
            \begin{tabular}{@{}lllr@{}}
            \toprule
            \textbf{Transition} & \textbf{Steps} & \textbf{Centers} & \textbf{$N$} \\
            \midrule
            Secure Balkan$\to$Def.\ Line & 2--19 & 2--8 & 184 \\
            Def.\ Line$\to$Secure Balkan & 3--19 & 2--8 & 145 \\
            Pressure Serbia$\to$Secure Balkan & 1--19 & 3--6 & 61 \\
            Def.\ Line$\to$Fortify Fronts & 3--19 & 3--8 & 30 \\
            \bottomrule
            \end{tabular}}
            \\[4pt]

            \smallbluebold{Pattern: } The top two transitions are inverses, creating a ``ping-pong'' loop (Balkan$\to$Def.$\to$Balkan$\to\!\cdots$) across \textit{all} step and center ranges with no consistent trigger.
            Confidence scores are uniformly low (0.23--0.24)---the retriever cannot tell when each skill applies.
            POSITION, DEFEND, and ATTACK all appear across the full center range (2--8), with no specialization.
            \end{minipage}
            \\[6pt]

            \noindent\colorbox{Blue!10}{\parbox{\dimexpr\linewidth-2\fboxsep}{\centering\small\textbf{Causal Mechanism: Skills $\to$ Temporal Structure $\to$ Action Diversity $\to$ Better Outcomes}}}
            \\[4pt]

            {\scriptsize
            \begin{tabular}{@{}lccl@{}}
            \toprule
            \textbf{Factor} & \textbf{Success ($c \geq 5$)} & \textbf{Failure ($c \leq 3$)} & \textbf{Mechanism} \\
            \midrule
            Action repeat rate & 22\% & 40\% & Fewer repeats $=$ more action-space exploration \\
            Action \#1 selection & 82\% & 90\% & Failures nearly degenerate---always pick first option \\
            Skill transitions & 2--3 & 2--3 & Same temporal schedule fires for both \\
            First center gain & Step 3--9 & Never & Early growth compounds; stagnation is self-reinforcing \\
            \bottomrule
            \end{tabular}}
            \\[4pt]

            \smallbluebold{Why skills help: }
            \textbf{(i)} They impose \textit{temporal structure}: the EXPLORE$\to$SETUP boundary at step~5 is a hard ``stop observing, start acting'' cutoff that breaks support loops.
            \textbf{(ii)} SETUP \textit{shifts the action distribution}: the subgoal prompt changes from ``scout intentions'' to ``secure supply centers,'' which is enough to move past purely defensive orders.
            \textbf{(iii)} Skills provide a \textit{floor, not a ceiling}---they prevent collapse (min${=}3$ vs.\ GPT-5.4 min${=}1$) even with poor action selection, because the phased structure stops the agent from abandoning centers.

        \end{minipage}
    };
    \end{tikzpicture}%
    \caption{\textbf{Skill retrieval patterns and causal mechanism in Diplomacy.}
    Skills function as a curriculum schedule for action exploration, as they impose temporal structure that broadens the action distribution and establishes a safety floor.}
    \label{fig:diplomacy-mechanism}
    \end{center}
\end{figure*}

To complement the aggregate results, we present step-level case studies in Candy Crush and Diplomacy that show how the evolved skill bank changes trajectory-level decisions across stages of play (Figures~\ref{candy-crush-comparison}--\ref{fig:diplomacy-comparison-cont}). 
We include strategy annotations and brief justifications explaining why \ours{} outperforms GPT-5.4 at key steps.
We also provide failure analysis for both GPT-5.4 and \ours{} in Figure~\ref{fig:diplomacy-failure}, highlighting qualitative differences and key insights. 
Finally, we examine skill retrieval and its causal role in Diplomacy in Figure~\ref{fig:diplomacy-mechanism}, showing how retrieved skills shape phase transitions and subsequent actions.

\newpage

\begin{figure}[!p]
    \begin{center}
    \begin{tikzpicture}
    \node [mybox,title=Candy Crush Strategy Comparison (cont'd)] (box){%
        \begin{minipage}{0.95\textwidth}
            \footnotesize
            \raggedright

            \noindent\colorbox{Blue!10}{\parbox{\dimexpr\linewidth-2\fboxsep}{\centering\small\textbf{Mid Game (Steps 11--34)}}}
            \\[4pt]

            \noindent
            \begin{minipage}[t]{0.48\linewidth}
            {\centering\textbf{\large GPT-5.4}\par}\vspace{0.3em}
            \smallbluebold{Step 23 -- Reactive midgame: }
            Still in \textit{[CLEAR]}, following the generic ``immediate match and cascade setup'' approach. \\
            \smallbluebold{Reward: } +6, cumul.=277 \\
            \smallbluebold{Strategy: } Still in reactive mode since step~22; no restructuring to set up step~28's cascade.

            \noindent\rule{\linewidth}{0.4pt}\vspace{0.3em}

            \smallbluebold{Step 28 -- Missed jackpot: }
            \textit{[CLEAR]} ``Complete the bottom-right candy match'' ((6,6),(6,7)). \\
            \smallbluebold{Reward: } +3, cumul.=327 \\
            \smallbluebold{Strategy: } Picks a corner match for 3 points; a higher-value cascade is available elsewhere on the board.
            \end{minipage}\hfill
            \begin{minipage}[t]{0.48\linewidth}
            {\centering\textbf{\large \ours{}}\par}\vspace{0.3em}
            \smallbluebold{Step 23 -- OPTIMIZE cycle: }
            First step of the mid-game OPTIMIZE phase. State: \texttt{note=Swapped G/P to clear threat, +11}. \\
            The previous CLEAR phase earned a steady 10--31 per step; the agent now switches back to preparation, setting up step~28's cascade. \\
            \smallbluebold{Reward: } +25, cumul.=355 (vs.\ 277) \\
            \smallbluebold{Why better: } Back in preparation while GPT-5.4 stays in CLEAR. Lead: +28\%.

            \noindent\rule{\linewidth}{0.4pt}\vspace{0.3em}

            \smallbluebold{Step 28 -- Cascade jackpot: }
            \textit{[OPTIMIZE]} 6th step of this phase. State: \texttt{note=Three new pairs created, threat of chain reaction}. \\
            The previous five OPTIMIZE steps yielded modest 7--24 rewards while patiently restructuring the board. Now \#9$\rightarrow$((3,5),(4,5)) triggers the episode's largest payoff. \\
            \smallbluebold{Reward: } \textbf{+62}, cumul.=455 (vs.\ +3, cumul.=327) \\
            \smallbluebold{Why better: } Five restructuring steps turn the flagged ``chain reaction'' into a 62-point cascade---20$\times$ GPT-5.4's reward at this step. Gap: +39\%.
            \end{minipage}
            \\[6pt]

            \noindent\colorbox{Blue!10}{\parbox{\dimexpr\linewidth-2\fboxsep}{\centering\small\textbf{Late Game (Steps 35--49)}}}
            \\[4pt]

            \noindent
            \begin{minipage}[t]{0.48\linewidth}
            \smallbluebold{Step 40 -- Endgame panic: }
            Switches to \textit{[EXECUTE]}: ``cut pairs fast with 10 moves left.'' \\
            \smallbluebold{Reward: } +21, cumul.=455 \\
            \smallbluebold{Strategy: } Switches to EXECUTE under time pressure. Score: 434 vs.\ \ours{} 573.

            \noindent\rule{\linewidth}{0.4pt}\vspace{0.3em}

            \smallbluebold{Step 46 -- Broken intent: }
            \textit{[EXECUTE]} ``maximize endgame points.'' \\
            \smallbluebold{Reward: } +3, cumul.=509 \\
            \smallbluebold{Strategy: } States ``maximize endgame points'' but scores only +3.

            \noindent\rule{\linewidth}{0.4pt}\vspace{0.3em}

            \smallbluebold{Step 49 -- Final move: }
            \textit{[EXECUTE]} ``horizontal four-match.'' \\
            \smallbluebold{Reward: } +10, \textbf{final score=547}
            \end{minipage}\hfill
            \begin{minipage}[t]{0.48\linewidth}
            \smallbluebold{Step 40 -- Endgame OPTIMIZE: }
            The skill description adapts to ``maximize remaining moves'' (compared to the mid-game's ``maximize cascades''). State: \texttt{note=Created 3 new pairs}. \\
            \smallbluebold{Reward: } +24, cumul.=637 (vs.\ 455) \\
            \smallbluebold{Why better: } Score: 573 vs.\ 434 (+32\%) with the same 10 moves left. OPTIMIZE keeps preparing; GPT-5.4 switches to EXECUTE. Lead: +40\%.

            \noindent\rule{\linewidth}{0.4pt}\vspace{0.3em}

            \smallbluebold{Step 46 -- CLEAR + \textsc{urgency}: }
            The system injects an urgency signal: ``very few moves left---maximise every action.'' Preceding steps scored +26, +21, +26. \\
            \smallbluebold{Reward: } +30, cumul.=760 (vs.\ +3, cumul.=509) \\
            \smallbluebold{Why better: } With the \textsc{urgency} signal, the agent sustains 21--30 points per step; GPT-5.4 scores +3 despite its own ``maximize endgame'' label.

            \noindent\rule{\linewidth}{0.4pt}\vspace{0.3em}

            \smallbluebold{Step 49 -- Final move: }
            \textit{[CLEAR]} + \textsc{urgency}. \#9$\rightarrow$((4,4),(5,4)). \\
            \smallbluebold{Reward: } +22, \textbf{final score=806} (vs.\ +10, 547). \textbf{+47\% improvement.}
            \end{minipage}

        \end{minipage}
    };
    \end{tikzpicture}%
    \caption{\textbf{Step-level comparison between GPT-5.4 and our method in Candy Crush} (2/2: Mid \& Late Game).
    See Figure~\ref{candy-crush-comparison} for setting and early game.}
    \label{candy-crush-comparison-cont}
    \end{center}
\end{figure}
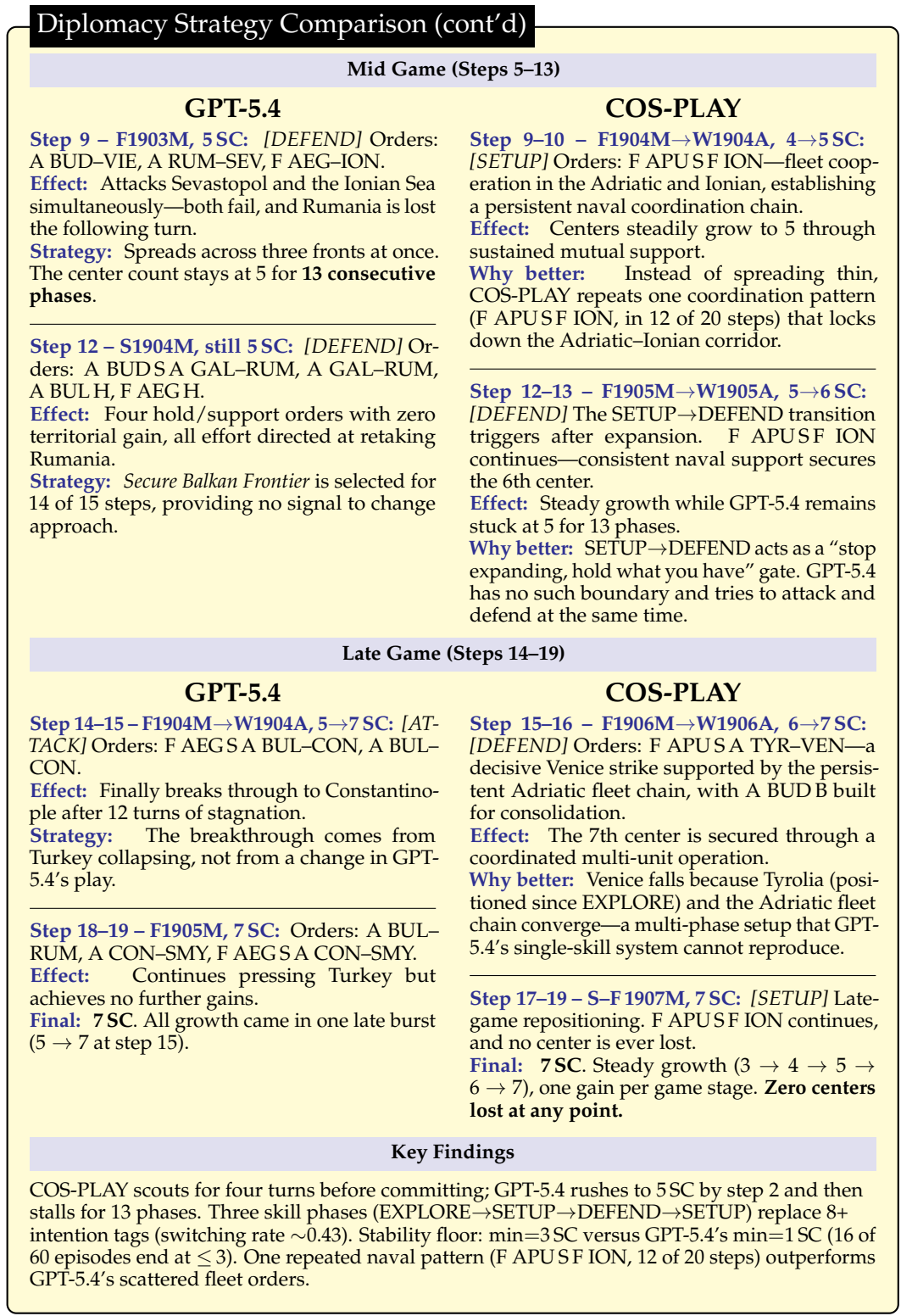

\input{showcases/showcase_diplomacy}

\begin{figure*}[!p]
    \begin{center}
    \begin{tikzpicture}
    \node [mybox,title={Diplomacy Failure Analysis: Stagnation vs.\ Collapse}] (box){%
        \begin{minipage}{0.95\textwidth}
            \footnotesize
            \raggedright

            \noindent\colorbox{Blue!10}{\parbox{\dimexpr\linewidth-2\fboxsep}{\centering\small\textbf{\ours{} Failure Mode: ``Stuck in Support Loop'' (5/28 episodes end at 3\,SC)}}}
            \\[4pt]

            {\scriptsize
            \begin{tabular}{@{}llll@{}}
            \toprule
            \textbf{Episode} & \textbf{Power} & \textbf{Most-Repeated Action} & \textbf{Repeat Rate} \\
            \midrule
            \texttt{diplomacy\_c40f3dc3} & Italy & F NAP\,S\,A ROM \;\;(14/20 steps) & 42\% \\
            \texttt{diplomacy\_2b255046} & Italy & F NAP\,S\,A ROM \;\;(15/20 steps) & 47\% \\
            \texttt{diplomacy\_95bd25f4} & Italy & A APU\,H (9), F NAP\,S\,A ROM (7) & 53\% \\
            \texttt{diplomacy\_e66ffd8e} & Austria & F ALB\,S\,A VEN--TRI \;\;(13/20 steps) & 37\% \\
            \texttt{diplomacy\_c5f6577c} & Germany & A BER--MUN (5), F BAL\,S\,A RUH--KIE (5) & 21\% \\
            \bottomrule
            \end{tabular}}
            \\[4pt]

            \smallbluebold{Root Cause: } The model has a strong \textbf{action-1 bias}: action~\#1 is selected in 85\% of all steps (90\% in failure episodes vs.\ 82\% in successes).
            Action~1 is usually a SUPPORT order, so failing episodes get stuck supporting the same unit over and over.
            Skill transitions still fire (EXPLORE$\to$SETUP$\to$DEFEND), but the action adapter does not vary its orders enough to break out.
            \\[6pt]

            \noindent\colorbox{Blue!10}{\parbox{\dimexpr\linewidth-2\fboxsep}{\centering\small\textbf{GPT-5.4 Failure Mode: ``Collapse'' (16/60 episodes, 27\%, end with $\leq$\,3\,SC)}}}
            \\[4pt]

            {\scriptsize
            \begin{tabular}{@{}lll@{}}
            \toprule
            \textbf{Episode} & \textbf{Center Trajectory} & \textbf{Failure Pattern} \\
            \midrule
            \texttt{episode\_010} & $3{\to}5{\to}5{\to}5{\to}4{\to}4{\to}2{\to}2{\to}1{\to}1$ & Peaked step~2, continuous decline for 17 steps \\
            \texttt{episode\_028} & $3{\to}5{\to}5{\to}5{\to}5{\to}3{\to}3{\to}3{\to}2{\to}2$ & Peaked step~2, lost 3 centers mid-game \\
            \texttt{episode\_043} & $3{\to}4{\to}4{\to}4{\to}4{\to}4{\to}3{\to}3{\to}2{\to}2$ & Slow bleed from mid-game onward \\
            \bottomrule
            \end{tabular}}
            \\[4pt]

            \smallbluebold{Root Cause: } \textit{Secure Defensive Line} stays active for 12--15 consecutive steps, with the intention cycling SURVIVE$\to$DEFEND$\to$SURVIVE.
            Centers keep dropping, but the skill bank has \textbf{no recovery skill} for losing positions.
            Once the decline starts, there is no way to pivot---the defensive skill orders retreats and disbands that accelerate the loss.
            \\[6pt]

            \noindent\colorbox{Blue!10}{\parbox{\dimexpr\linewidth-2\fboxsep}{\centering\small\textbf{Key Insight \& Qualitative Outplay}}}
            \\[4pt]

            \noindent
            \begin{minipage}[t]{0.48\linewidth}
            \smallbluebold{Stagnation vs.\ Collapse: }
            \ours{} fails by \textbf{stagnation}---stuck at 3\,SC, never growing.
            GPT-5.4 fails by \textbf{collapse}---grows to 4--5\,SC, then drops to 1--2.
            Stagnation is the safer failure mode: the agent never loses its starting centers.
            The phased skill structure provides a \textit{floor}---even with poor action selection, the agent does not abandon centers
            (min${=}3$ vs.\ GPT-5.4 min${=}1$).
            \end{minipage}\hfill
            \begin{minipage}[t]{0.48\linewidth}
            \smallbluebold{Outplay Example (France, Step 8, F1904M): }
            France holds only 3\,SC but has units in aggressive positions: A~TYR (deep in Austrian territory), A~BEL, F~MAO.
            \ours{} plays A~BEL\,S\,F~NTH--HOL---a joint strike with England---taking \textbf{3 new centers in one turn} ($3 \to 6$).
            The preceding 8-turn ``holding pattern'' (BRE$\leftrightarrow$MAO cycling) made GPT-5-mini opponents commit elsewhere; the simultaneous strike went unanswered.
            At the same stage, GPT-5.4's best Austria episode sits at 5\,SC with four hold/support orders and \textit{Secure Balkan Frontier} selected for 14 of 15 steps.
            \end{minipage}

        \end{minipage}
    };
    \end{tikzpicture}%
    \caption{\textbf{Failure analysis for both methods in Diplomacy.}
    \ours{} fails by stagnation (5/28 episodes plateau at 3\,SC); GPT-5.4 fails by collapse (16/60 decline to 1--2\,SC). Stagnation is the safer failure mode.}
    \label{fig:diplomacy-failure}
    \end{center}
\end{figure*}

\end{document}